%% file: textoPrincipal.tex
\documentclass{SBCbookchapter}
\usepackage[T1]{fontenc}
\usepackage[utf8]{inputenc}

\usepackage{listings}
\usepackage[brazil]{babel}
\usepackage{graphicx}
\usepackage{multirow}
\usepackage{blindtext}
\usepackage{spverbatim}

\usepackage{xcolor}
\usepackage{wrapfig}
\usepackage{comment}
\usepackage{url}
\usepackage{hyperref}
\definecolor{codegreen}{rgb}{0,0.6,0}
\definecolor{codegray}{rgb}{0.5,0.5,0.5}
\definecolor{codepurple}{rgb}{0.58,0,0.82}
\definecolor{backcolour}{rgb}{0.95,0.95,0.92}

\lstdefinestyle{mystyle}{
    backgroundcolor=\color{backcolour},   
    commentstyle=\color{codegreen},
    keywordstyle=\color{magenta},
    numberstyle=\tiny\color{codegray},
    stringstyle=\color{codepurple},
    basicstyle=\ttfamily\footnotesize,
    breakatwhitespace=false,         
    breaklines=true,                 
    captionpos=b,                    
    keepspaces=true,                 
    numbers=left,                    
    numbersep=5pt,                  
    showspaces=false,                
    showstringspaces=false,
    showtabs=false,                  
    tabsize=2
}

\lstdefinelanguage{json}{
    basicstyle=\ttfamily\scriptsize,
    numbers=left,
    numberstyle=\tiny,
    stepnumber=1,
    numbersep=8pt,
    showstringspaces=false,
    breaklines=true,
    frame=single,
    string=[s]{"}{"},
    stringstyle=\color{blue},
    comment=[l]{//},
    commentstyle=\color{gray},
    morecomment=[l]{\#},
    keywordstyle=\color{purple},
    keywords={true,false,null}
}

\usepackage{tikz}

\usetikzlibrary{matrix,positioning,arrows,backgrounds}
\usetikzlibrary{arrows.meta}
\usetikzlibrary{decorations.pathreplacing,calc}
\usetikzlibrary{fit}
\usetikzlibrary{decorations.markings}
\usetikzlibrary{chains,scopes,positioning,backgrounds,shapes,fit,shadows,calc}
\usepackage{pgfplots}
\pgfplotsset{compat=1.17}
\usepackage{pgfplotstable}
\usepgfplotslibrary{fillbetween}
\usepackage{siunitx}
\usetikzlibrary{patterns}
\usetikzlibrary{shapes.geometric}
\usepackage{multirow}
\usepackage[table,xcdraw]{xcolor}
\usetikzlibrary{calc,shapes.callouts,shapes.arrows}
\usetikzlibrary{arrows,shadows,positioning}
\usetikzlibrary{arrows.meta}
\usepackage{ragged2e}

\usepackage{subfiles}
\usepackage{minted}
\usepackage{caption}
\usepackage{xcolor}
\usepackage{amsmath}
\usepackage{amssymb}
\usepackage{cleveref}
\usepackage{listings}
\setminted{
    fontsize=\footnotesize,
    breaklines=true,
    breakanywhere=true, 
    linenos=true,
    frame=single,
    framesep=2mm,
    baselinestretch=1.2,
    bgcolor=gray!5,
    tabsize=4
}

\newcommand\VLMos{GPT (OS)}
\newcommand\VLMoai{GPT (OAI)}
\newcommand\VLMds{DeepSeek}

\title{Grandes Modelos de Linguagem Multimodais (MLLMs): Da Teoria à Prática}

\author{Neemias da Silva\inst{1}, Júlio C. W. Scholz\inst{1}, John Harrison\inst{1}, Marina Borges\inst{1}, Paulo\\ Ávila\inst{1}, Frances A Santos\inst{2}, Myriam Delgado\inst{1}, Rodrigo Minetto\inst{1}, Thiago H Silva\inst{1,3}}

\address{Universidade Tecnológica Federal do Paraná (UTFPR) - Curitiba, Brasil
\nextinstitute
 Universidade Estadual de Campinas (UNICAMP), Campinas, Brasil
\nextinstitute
  University of Toronto, Toronto, Canada
  \email{neemiasbuceli@alunos.utfpr.edu.br, \{myriamdelg,rminetto\}@utfpr.edu.br,}
 \email{th.silva@utoronto.ca}}

\begin{document}
\maketitle

\sloppy

\begin{abstract}
Multimodal Large Language Models (MLLMs) combine the natural language understanding and generation capabilities of LLMs with perception skills in modalities such as image and audio, representing a key advancement in contemporary AI. This chapter presents the main fundamentals of MLLMs and emblematic models. Practical techniques for preprocessing, prompt engineering, and building multimodal pipelines with LangChain and LangGraph are also explored. For further practical study, supplementary material is publicly available online: \url{https://github.com/neemiasbsilva/MLLMs-Teoria-e-Pratica}. Finally, the chapter discusses the challenges and highlights promising trends.
\end{abstract}

\begin{resumo}
\begin{otherlanguage}{brazilian}
Os Grandes Modelos de Linguagem Multimodais (MLLMs do inglês \textit{Multimodal Large Language Models}) combinam a capacidade de compreensão e geração de linguagem natural dos LLMs com habilidades de percepção em modalidades como imagem e áudio, constituindo um avanço central na IA contemporânea. O capítulo apresenta os principais fundamentos dos MLLMs e modelos emblemáticos. Também são exploradas técnicas práticas de pré-processamento, engenharia de prompts e construção de pipelines multimodais com LangChain e LangGraph. Para aprofundamento prático, material complementar está disponível publicamente na web: \url{https://github.com/neemiasbsilva/MLLMs-Teoria-e-Pratica}. Por fim, o capítulo discute desafios e aponta tendências promissoras.
\end{otherlanguage}
\end{resumo}

\section{Introdução e Contextualização}\label{sec:introducao}

Os Grandes Modelos de Linguagem Multimodais (MLLMs do inglês \textit{Multimodal Large Language Models}) representam um dos avanços mais significativos na interseção entre inteligência artificial, processamento de linguagem natural e visão computacional \cite{caffagni2024revolution,wu2023multimodal,yin2024survey}. Diferentemente de seus predecessores - os Grandes Modelos de Linguagem (LLMs do inglês \textit{Large Language Models}) que operavam predominantemente com texto - os MLLMs possuem a capacidade notável de processar, interpretar e gerar conteúdo a partir de múltiplas modalidades, como texto, imagem, áudio e, em alguns casos, vídeo. 

\begin{figure}[h!]
    \centering
    \includegraphics[width=0.7\textwidth]{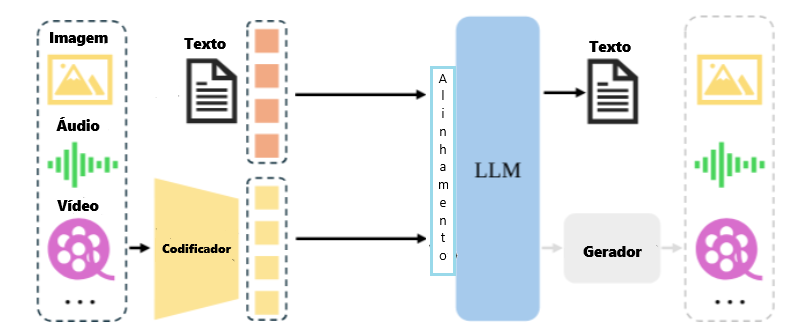} 
    \caption{Visão geral da arquitetura de um MLLM: entradas de diferentes modalidades são codificadas e alinhadas à entrada textual para serem processadas de forma conjunta por um LLM. A saída do MLLM pode ser textual ou multimodal (via gerador específico). Figura adaptada de \cite{yin2024survey}.}
 \label{fig:MLLMoverview}
\end{figure}

 A Figura  \ref{fig:MLLMoverview} traz uma visão geral da arquitetura de um MLLM. Os dados de entrada em múltiplas modalidades são primeiro codificados por modelos especializados e, assim como texto, são transformados em  representações vetoriais chamadas \textit{embeddings}. Um componente fundamental de um MLLM é o módulo de alinhamento, que adapta os  \textit{embeddings}  ao formato compreensível pelo LLM (núcleo em azul formado em geral por um \textit{decoder} de \textit{Transformer} - o mesmo componente arquitetural que impulsiona LLMs de texto puro como o GPT-4 e o LLaMA). Este então processa os dados em conjunto, relacionando texto e outras modalidades. Assim, um MLLM consegue descrever, responder e raciocinar sobre diferentes tipos de informação. Na saída, há modelos que geram apenas texto e outros que, a partir de um mesmo processo de raciocínio interno, utilizam um gerador para converter \textit{tokens} gerados pelo LLM em múltiplas modalidades.

Os MLMMs representam, portanto, um aprimoramento dos modelos de linguagem e visão - VLLMs do inglês  \textit{Vision-Language Large Models} - que trabalham apenas com texto e imagem. Nesses modelos, tipicamente, um codificador de visão projeta as imagens em um espaço latente e um adaptador de \textit{prompt} alinha essa representação com a entrada do LLM.

\subsection{Relevância dos MLLMs na Web e Mídias Digitais}

 A \textit{Web} moderna é um ecossistema essencialmente multimodal. Por exemplo, plataformas de \textit{e-commerce} dependem da combinação de imagens de produtos com descrições textuais, redes sociais como \textit{Instagram} e \textit{TikTok} são construídas sobre a interação entre vídeo, áudio, imagem e texto, e até mesmo a busca na \textit{web} está evoluindo para permitir consultas por imagem. Os MLLMs surgem como a ferramenta ideal para navegação, organização, moderação e criação de conteúdo para este ambiente complexo. Eles permitem diferentes recursos.

    \textbf{Busca e recuperação semântica}: Encontrar produtos, notícias ou conteúdo multimodal com base em consultas complexas que envolvem tanto texto quanto imagem.
    
    \textbf{Acessibilidade}: Gerar descrições textuais automáticas para imagens (\textit{alt text}), tornando a web mais acessível para usuários com limitação visual.
   
    \textbf{Criação de conteúdo}: Auxiliar na criação de conteúdos variados, como gerar legendas para \textit{posts} em redes sociais, criar ilustrações a partir de rascunhos textuais ou mesmo produzir vídeos curtos baseados em um roteiro.
   
    \textbf{Moderação de conteúdo}: Identificar conteúdo inadequado de forma mais contextual, analisando simultaneamente imagens e o texto associado, o que é mais eficaz do que analisar cada modalidade isoladamente.

\begin{figure}[htbp]
    \centering
    \includegraphics[width=0.85\textwidth]{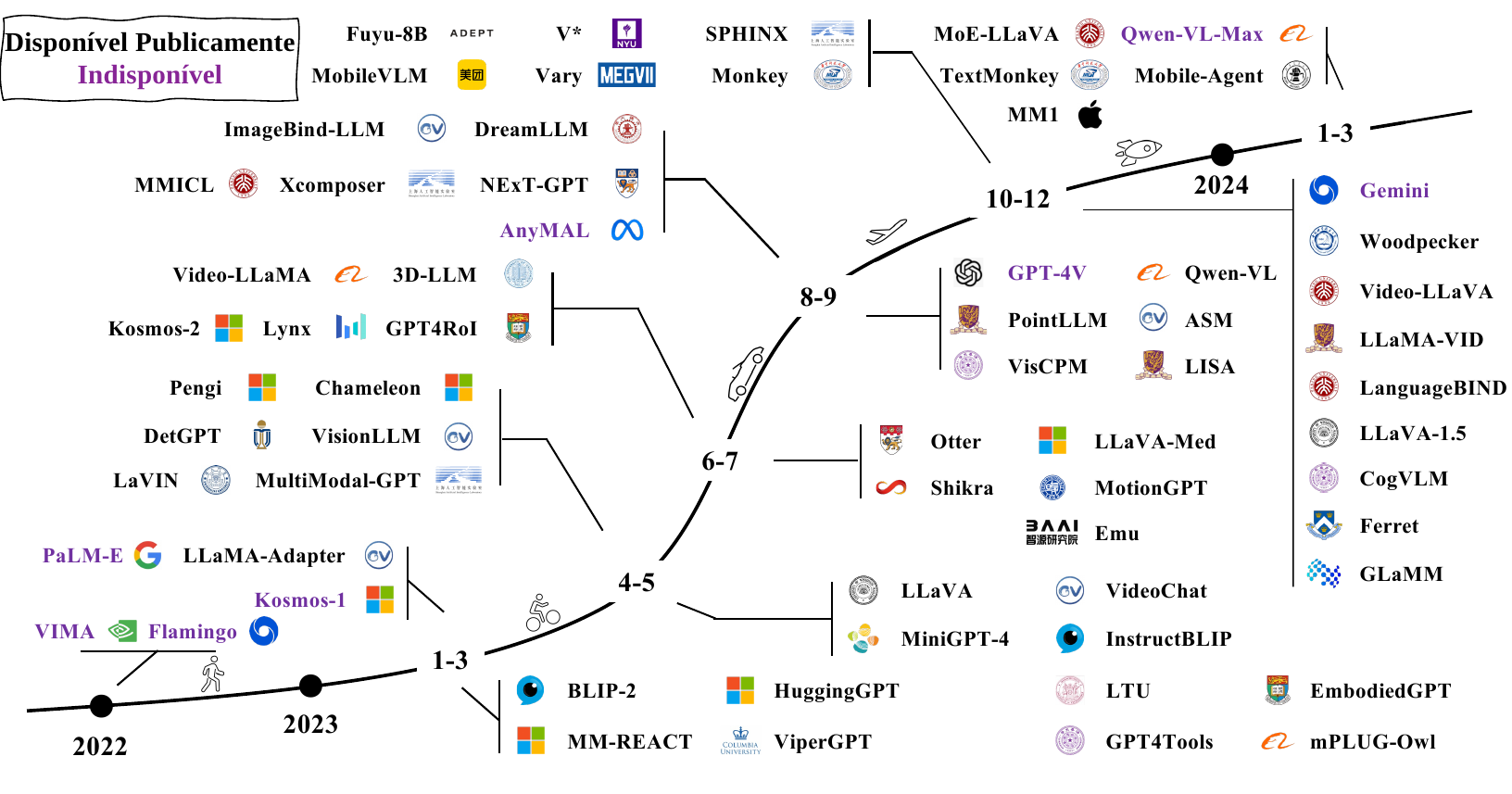} 
    \caption[Linha do tempo de MLLMs emblemáticos]{Linha do tempo simplificada ilustrando a evolução de alguns MLLMs emblemáticos, mostrando a diversificação entre modelos de código aberto e fechado (indisponível publicamente). Figura adaptada de \cite{yin2024survey}.}
    \label{fig:timeline_mllms}
\end{figure}

A Figura \ref{fig:timeline_mllms} ilustra a rápida evolução e proliferação de MLLMs nos últimos anos, destacando a aceleração no desenvolvimento de modelos tanto proprietários quanto de código aberto, um testemunho do intenso interesse e investimento nesta área.

\subsection{Desafios e Oportunidades no Processamento Conjunto das Multimodalidades}

A motivação para o desenvolvimento e estudo dos MLLMs reside na combinação de desafios complexos e oportunidades transformadoras provenientes do processamento conjunto de diferentes modalidades. 
Os desafios são significativos \cite{liu2025unveiling,kalai2025allucinate}. Diferentes modalidades existem em espaços de representação fundamentalmente distintos, por exemplo, \textit{pixels} para imagens e \textit{tokens} para texto. O principal obstáculo é, portanto, a criação de uma ``ponte'' semântica entre esses espaços, um processo conhecido como alinhamento de \emph{embeddings}. Outros desafios críticos incluem a dessincronização de modalidades, onde o texto associado a uma imagem pode ser ambíguo, enganoso ou descrever apenas parcialmente o conteúdo visual; as alucinações multimodais, onde os MLLMs podem inventar detalhes visuais ou atributos textuais não presentes nas entradas, gerando informações incorretas com alta confiança; o custo computacional elevado do treinamento e inferência com dados de alta dimensionalidade como imagens, levantando questões sobre eficiência energética e acessibilidade no desenvolvimento e no uso desses modelos; e as questões de viés e justiça, já que os MLLMs podem herdar e até amplificar vieses presentes nos dados de treinamento multimodais, levando a estereótipos prejudiciais em suas saídas.

No entanto, apesar dos desafios, os MLLMs atuais possibilitam várias aplicações em diversas áreas \cite{kuang2025natural,da2025multimodal}. A capacidade de ``raciocinar'' sobre o mundo de forma integrada, tal como os humanos fazem naturalmente, abre um leque de aplicações anteriormente difíceis (ou não viáveis) de serem desenvolvidas. Por exemplo, MLLMs viabilizam o desenvolvimento de assistentes inteligentes avançados que podem verdadeiramente ``ver'' o mundo através da câmera de um \textit{smartphone} e interagir com ele de forma contextual, ajudando usuários desde a identificação de um organismo vegetal até a solução de problemas em manuais de instrução. Na área de educação personalizada, permitem a criação de tutores interativos que utilizam diagramas, ilustrações e textos para explicar conceitos complexos de forma adaptativa. Para análise de dados complexos, possibilitam a interpretação automatizada de relatórios médicos que combinam imagens de raio-X com laudos textuais, ou análise de imagens de satélite acopladas a dados geolocalizados para monitoramento ambiental. Na automação robótica, fornecem a robôs uma compreensão de alto nível de seu ambiente, permitindo que executem tarefas baseadas em instruções naturais como ``pegue a xícara vermelha sobre a mesa'' (compreensão exemplificada por modelos como o \textit{PaLM-E} \cite{driess2023palmeembodiedmultimodallanguage}).

\subsection{Visão Geral do Material}

Este texto foi concebido para guiar os interessados em uma jornada compreensiva pelos Grandes Modelos de Linguagem Multimodais, partindo dos seus fundamentos teóricos até a implementação prática. A estrutura foi planejada para oferecer uma base sólida a iniciantes, ao mesmo tempo que apresenta tópicos avançados e o estado da arte para praticantes experientes. O conteúdo está organizado da seguinte forma:

\begin{itemize}
    \item \textbf{Seção \ref{sec:fundamentos}: Fundamentos de LLMs e Multimodalidade} - Esta seção estabelece as bases, revisitando a arquitetura \textit{Transformer} e a evolução para os LLMs modernos baseados apenas em \textit{decoders}. Em seguida, aborda-se uma parte chave da multimodalidade, explicando o conceito de alinhamento de \emph{embeddings} e das principais arquiteturas de MLLMs, bem como uma análise de modelos emblemáticos como GPT-4V, Gemini e LLaVA.
    
    \item \textbf{Seção \ref{sec:pipeline}: Técnicas para Pipeline Multimodal} - Nesta seção são apresentadas, de forma prática,  estratégias para o pré-processamento de dados multimodais e técnicas avançadas de engenharia de \textit{prompt}. A seção inclui ainda o uso de arcabouços como LangChain e LangGraph para construir \textit{pipelines} robustos e escaláveis que integram MLLMs a outras ferramentas.
    
    \item \textbf{Seção \ref{secAppPraticas}: Aplicações Práticas e Casos de Uso} - Esta seção apresenta exemplos concretos de aplicação de MLLMs, incluindo geração de descrições de imagens, perguntas e respostas multimodais (VQA do inglês \textit{Visual Question Answering}) e uma análise detalhada de um estudo de caso real: o arcabouço MLLMsent para análise de sentimentos em imagens.
    
    \item \textbf{Seção \ref{secHandsON}: \textit{Hands-on}: Tutorial Prático} - Nesta seção, utilizando modelos acessíveis, \textit{notebooks} práticos mostram desde a classificação direta de imagens com MLLMs até a construção de um \textit{pipeline} completo de classificação via descrições geradas.
    
    \item \textbf{Seção \ref{secLimitacoes}: Limitações e Oportunidades} - Esta seção discute, de forma crítica, as principais limitações atuais dos MLLMs, como alucinações, viés e custo computacional; também explora as tendências promissoras para pesquisas futuras, incluindo modelos menores e mais eficientes e o papel dos agentes multimodais.

    \item \textbf{Seção \ref{secConclusao}: Conclusões}.
\end{itemize} 

Este material visa fornecer ao interessado não apenas uma compreensão teórica 
do funcionamento dos MLLMs, mas também as habilidades práticas necessárias para  desenvolver e integrar essas tecnologias transformadoras em seus próprios projetos, contribuindo para a próxima geração de aplicações inteligentes na Web e nas mídias digitais.

\section{Fundamentos de LLMs e Multimodalidade }
\label{sec:fundamentos}

Esta seção apresenta uma visão condensada das arquiteturas que sustentam LLMs e MLLMs. A Subseção~\ref{subsec:fundamentos_arquitetura} revisa a evolução dos \textit{Transformers} aos LLMs modernos; na Subseção~\ref{subsec:ponte_multimodalidade} são descritos os mecanismos de alinhamento de \textit{embeddings} que viabilizam a multimodalidade; já a  Subseção~\ref{subsec:arquiteturas_mllms} detalha como diferentes arquiteturas de MLLMs incorporam múltiplas modalidades; e, por fim, a Subseção~\ref{subsec:modelos_emblematicos} apresenta modelos e arcabouços representativos.

\subsection{Arquiteturas Base: Dos \textit{Transformers} aos LLMs Modernos}
\label{subsec:fundamentos_arquitetura}

A compreensão dos MLLMs requer uma fundamentação sobre a arquitetura que os sustenta. No centro dessa evolução está o mecanismo de atenção e a arquitetura \textit{Transformer}.

Introduzido pelo artigo seminal \textit{``Attention is All You Need''} \cite{vaswani2017attention}, o mecanismo de auto-atenção (\textit{self-attention}) permite que um modelo pondere a importância relativa de diferentes \textit{tokens}\footnote{Um token é a unidade básica de processamento de texto em modelos de linguagem, podendo corresponder a uma palavra, subpalavra ou caractere.} em uma sequência textual, considerando dependência de longo alcance. Esse mecanismo é a base para codificar contexto de maneira eficiente.
O \textit{Transformer} original é composto por uma pilha
de \textit{encoders}, que mapeiam uma sequência de entrada para uma representação latente, e uma pilha de \textit{decoders}, que geram uma sequência de saída a partir dessa representação. Essa arquitetura inovadora abriu caminho para especializações em duas direções principais. 

Modelos baseados em \textit{encoder}, como o BERT \cite{devlin2019bert}, que são otimizados para tarefas de \textbf{compreensão de linguagem}. Usam atenção bidirecional, acessando simultaneamente o contexto à esquerda e à direita de cada token, o que os torna adequados para análise de texto, classificação e resposta a perguntas condicionadas a um contexto dado. São considerados modelos não autorregressivos (preveem os \textit{tokens} simultaneamente, sem dependência de \textit{tokens} anteriores).
Por outro lado, modelos baseados em \textit{decoder}, como a família GPT \cite{radford2018improving, radford2019language}, são otimizados para a \textbf{geração de linguagem}. Nesse caso, a atenção é causal (ou mascarada), acessando apenas o contexto anterior na sequência para prever o próximo \textit{token}. A geração ocorre de forma autorregressiva, \textit{token} por \textit{token}, o que garante fluidez na produção textual.

No uso contemporâneo, o termo LLMs refere-se predominantemente a \textit{decoders} autorregressivos, como as famílias GPT e LLaMA \cite{touvron2023llama}. Com centenas de bilhões de parâmetros treinados em corpora massivos de texto, esses modelos demonstram capacidades emergentes de ``raciocínio'', execução de instruções e geração de linguagem natural \cite{wei2022emerging}.

\begin{figure}[ht]
\centering
\includegraphics[width=\textwidth]{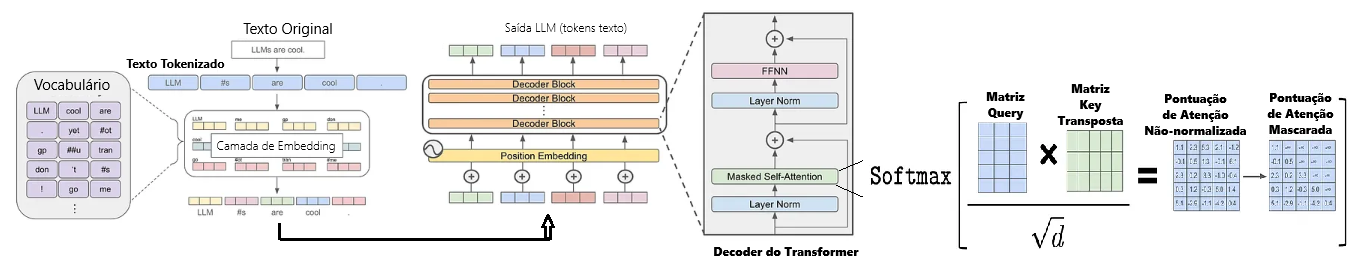}
\caption{Visão geral do funcionamento de um LLM. Figura adaptada de \cite{transf}.}
\label{fig:LLMs}
\end{figure}

A Figura \ref{fig:LLMs} ilustra o funcionamento geral de um LLM. Primeiro, a entrada textual é tokenizada, ou seja, dividida em unidades menores (\textit{tokens}). Cada \textit{token} é convertido em um vetor de representação numérica densa (\textit{embedding}), sendo somado ao com o vetor de representação numérica densa da posição de cada \textit{token} na sequência (\textit{positional encoding embedding}). Esses vetores são enviados a uma pilha de \textit{decoders} do \textit{Transformer}, onde ocorre o processamento autorregressivo 
gerando um \textit{token} de saída por vez. No \textit{decoder}, a etapa central é a auto-atenção mascarada, onde cada \textit{token} gera três vetores: $\mathbf{q}$ (\textit{query}),  $\mathbf{k}$ (\textit{key}) e  $\mathbf{v}$ (\textit{value}); os quais, agregados para todos os \textit{tokens} de entrada, formam as respectivas matrizes, $Q$, $K$ e $V$. Calcula-se a similaridade entre $Q$ e $K$, aplica-se \textit{softmax} e multiplica-se pelos valores $V$, mas com uma máscara de elementos com valores infinitos que impede o modelo de ver \textit{tokens} futuros. O resultado, somado a uma conexão residual, passa por  uma camada de normalização, garantindo estabilidade e aprendizado eficiente. Em seguida, os vetores passam por uma rede neural de propagação direta (FFNN do inglês \textit{feed-forward neural network}).
Esse processo realizado em cada bloco do \textit{decoder} se repete por todas as camadas do \textit{Transformer} até a última, onde uma camada linear projeta o vetor final para o vocabulário, produzindo uma distribuição de probabilidade para o próximo \textit{token}. O modelo, então, escolhe esse \textit{token} e o reinsere no processo, gerando o texto de forma incremental.

A multimodalidade surge como uma extensão natural desses LLMs. Conforme mostrado na Figura~\ref{fig:MLLMoverview}, a inovação crucial reside na incorporação de \textit{encoders} 
 especializados que convertem entradas não-textuais (como \textit{pixels} de imagem) em representações (\textit{embeddings}) que podem ser projetadas no espaço latente dos LLMs. Através de técnicas de alinhamento multimodal, como ocorre no CLIP \cite{radford2021learning} (ver Figura~\ref{fig:clip}), o modelo aprende a associar conceitos  de diferentes modalidades, permitindo que este ``entenda'' uma imagem e converse sobre ela, ou que gere uma imagem a partir de uma descrição textual. 

Modelos como o GPT-4V \cite{openai2023gpt4v} e o Gemini \cite{gemini2023report} partem da mesma base de \textit{decoders} autorregressivos para textos, mas incorporam informações adicionais, como visão ou áudio. A estratégia predominante consiste em projetar \textit{embeddings} de todas as modalidades num espaço comum compreensível para o \textit{decoder}. Assim, o núcleo autorregressivo é mantido, mas agora gera respostas para entradas multimodais.
Essa herança arquitetural é o que permite aos MLLMs gerar linguagem de forma coerente sobre conteúdos além dos textuais, como visuais e auditivos.

\subsection{A Ponte para a Multimodalidade: Alinhamento de \textit{Embeddings}}
\label{subsec:ponte_multimodalidade}
 
Um ponto-chave dos MLLMs é a capacidade de integrar informações de modalidades distintas, como texto, imagem e áudio, em um modelo coerente. O desafio é que cada modalidade possui uma representação numérica nativa (\textit{tokens}, \textit{pixels}, sinais de áudio, respectivamente), que habita espaços vetoriais distintos e não comparáveis diretamente. A solução usual está no alinhamento de \textit{embeddings}.

Cada modalidade é inicialmente processada por uma arquitetura especializada. Modelos de linguagem baseados em \textit{Transformer}, como o GPT, mapeiam sequências de \textit{tokens} para um espaço vetorial \(\mathcal{T}\). \textit{Encoders} visuais, como \textit{Vision Transformers} (ViTs) \cite{dosovitskiy2020image}, projetam \textit{patches} de imagens em um espaço vetorial \(\mathcal{V}\). Modelos de áudio, como \textit{wav2vec 2.0} \cite{baevski2020wav2vec}, convertem  sinais de áudio  em um espaço \(\mathcal{A}\).
Sem um mecanismo de alinhamento, os espaços \(\mathcal{T}\), \(\mathcal{V}\) e \(\mathcal{A}\) são como línguas diferentes: um modelo de linguagem não pode compreender diretamente um vetor de imagem. O \textit{joint embedding} propõe justamente mapear essas modalidades em um espaço semântico latente compartilhado \(\mathcal{C}\), ou seja, uma linguagem universal que o modelo entende. Nesse espaço, conceitos semelhantes, como a palavra ``cachorro'', a foto de um cachorro e o som de um latido, são projetados em regiões vizinhas \cite{baltrusaitis2018multimodal}. Essa projeção permite que o LLM, operando em \(\mathcal{C}\), processe e gere conteúdo condicionado em múltiplas modalidades.

\begin{figure}[ht]
\centering
\includegraphics[width=.9\textwidth]{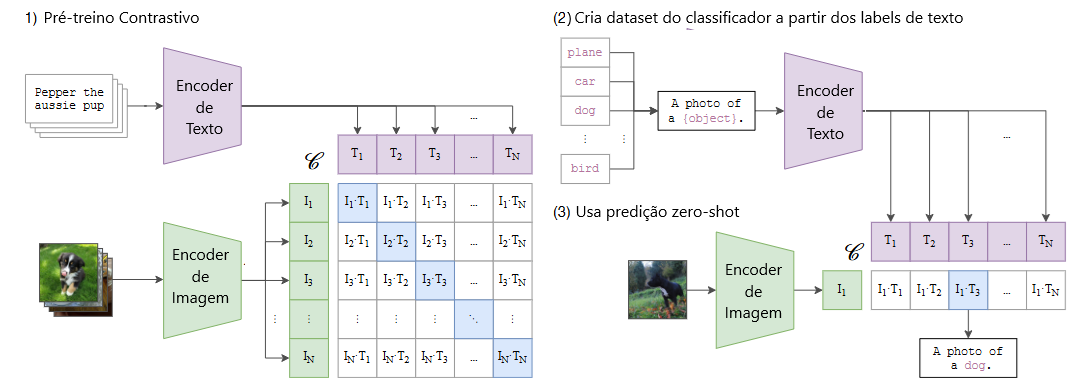}
\caption{Visão geral do CLIP. Figura adaptada de \cite{radford2021learning}.}
\label{fig:clip}
\end{figure}

Uma técnica conhecida para produzir esse alinhamento é o aprendizado contrastivo, no qual o modelo aprende representações úteis comparando exemplos relacionados (alinhados) e não relacionados~\cite{chen2020simple}. O CLIP (\textit{Contrastive Language–Image Pre-training}) \cite{radford2021learning} é um exemplo emblemático. Ele é treinado com milhões de pares (imagem, legenda) e, conforme ilustrado na Figure~\ref{fig:clip}, utiliza dois \textit{encoders}: um \textit{encoder} de texto (ex.: \textit{Transformer}) para mapeamento em \(\mathcal{T}\) e um \textit{encoder} de imagem (ex.: ViT) para mapeamento em \(\mathcal{V}\). No CLIP, os \textit{encoders} visual e textual são treinados conjuntamente para que ambos produzam  \textit{embeddings} em um espaço latente compartilhado \(\mathcal{C}\), e o objetivo do aprendizado  é maximizar a similaridade entre pares  que são relacionados (diagonal principal) e minimizá-la para pares não relacionados (fora da diagonal principal).
O resultado é um espaço multimodal compartilhado \(\mathcal{C}\) que permite tarefas de recuperação \textit{zero-shot} (inferência em um contexto novo, diferente do treino).

Nos MLLMs, o CLIP atua como uma ponte para a incorporação de imagens, fornecendo um espaço compartilhado \(\mathcal{C}\) no qual representações visuais e textuais são alinhadas de forma simétrica \cite{radford2021learning}. Durante o treinamento do CLIP, um par de \textit{encoders}, um visual e outro textual, é otimizado conjuntamente para ter seus \textit{embeddings} de imagens e descrições projetados em um mesmo espaço latente, de modo que pares correspondentes fiquem próximos e pares não correspondentes se afastem.
Em modelos multimodais baseados em LLMs, esses \textit{embeddings} visuais produzidos pelo CLIP podem ser posteriormente projetados para 
um espaço compreensível pelo modelo de linguagem~\cite{liu2023visual}, permitindo que o LLM trate a informação visual como uma forma de entrada textual compatível. Assim, o modelo não “traduz” a imagem diretamente para o texto, mas permite que o LLM interprete a informação visual como \textit{tokens} compatíveis com seu espaço latente.

\begin{figure}[!htb]
\centering
   \input{scheme-how-it-works}
\caption{Processo de construção de \textit{embeddings} visuais (a) e textuais (b) em modelos multimodais. Figura adaptada  de \cite{raschka2024}.}

\label{fig:overviewEmbed}
\end{figure}
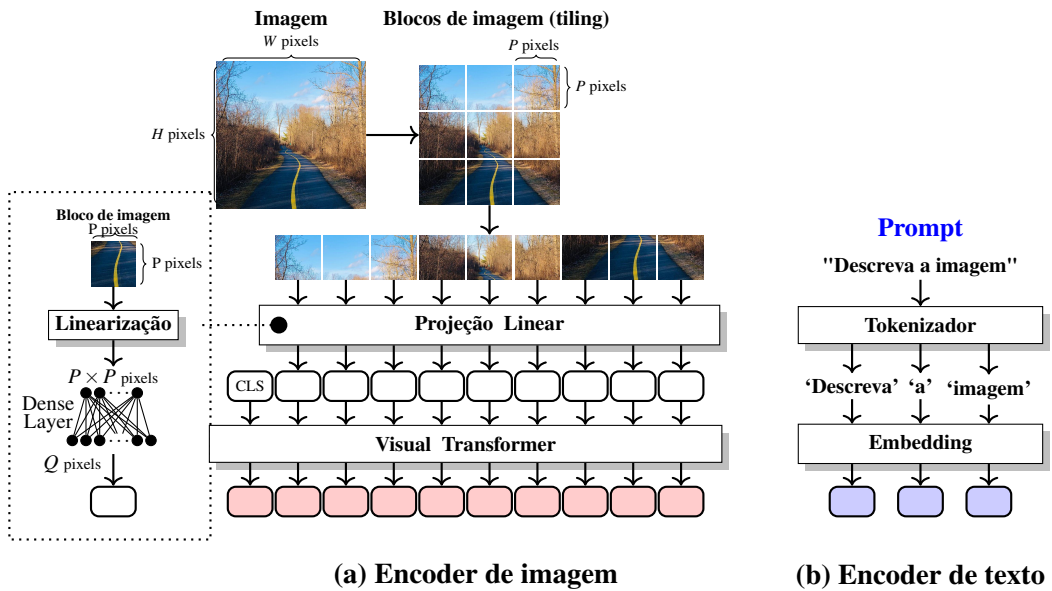

A Figura~\ref{fig:overviewEmbed} ilustra o processo de geração dos \textit{embeddings} de imagem e texto empregados como entrada em modelos de linguagem multimodais. 

Inicialmente, a imagem fornecidade pelo usuário com largura e altura de
$W \times H$ \textit{pixels} é dividida em pequenos blocos (ou \textit{patches}) de $P \times P$ \textit{pixels}, que são linearizados e empilhados sequencialmente. Cada \textit{patch} é então projetado em um espaço vetorial de $Q$ dimensões por meio de uma camada totalmente conectada, resultando em uma sequência de vetores que, juntamente com o \textit{token} especial [CLS], formam a entrada para um \textit{Vision Transformer} (ViT). O valor de $Q$ é escolhido de tal forma a se encaixar na entrada de cada \textit{token} definido pelo ViT. Após o processamento pelo ViT, são obtidos os \textit{image embeddings} correspondentes, que codificam a informação visual de maneira contextualizada. 
De forma análoga, a sequência textual, exemplificada pelo \textit{prompt} ``Descreva a imagem”,  é tokenizada, transformando cada palavra em uma unidade discreta. Esses são os \textit{tokens} de texto, cujos \textit{embeddings} são posteriormente projetados, assim como os \textit{embeddings} visuais, em um mesmo espaço latente. 

\subsection{Arquiteturas de MLLMs: Como a Multimodalidade é Incorporada}
\label{subsec:arquiteturas_mllms}

Com os fundamentos estabelecidos, é possível compreender as principais estratégias de construção de MLLMs. A ideia central é utilizar o LLM como um ``sistema cognitivo central'', ao qual se conectam \textit{encoders} especializados para processar outras modalidades. A questão é, portanto, de interface: como conectar essas representações ao modelo de linguagem de forma que ele possa ``raciocinar'' sobre informações multimodais. 

\begin{figure}[!htb]
   \input{scheme-modalities}
   \caption{Estratégias de fusão em MLLMs: (a) A fusão precoce integra \textit{tokens} de imagem e texto já na entrada do modelo, permitindo que o decodificador processe ambas as modalidades de forma unificada; (b) A fusão tardia incorpora os \textit{embeddings} visuais aos \textit{embeddings} textuais, em geral por meio do mecanismo de atenção cruzada multi cabeças (\textit{multi-head cross-attention}); (c) A fusão híbrida combina ambas as abordagens, fornecendo \textit{tokens} de imagem tanto na entrada quanto nas camadas de atenção.}
\label{fig:AlignmentSchemes}
\end{figure}
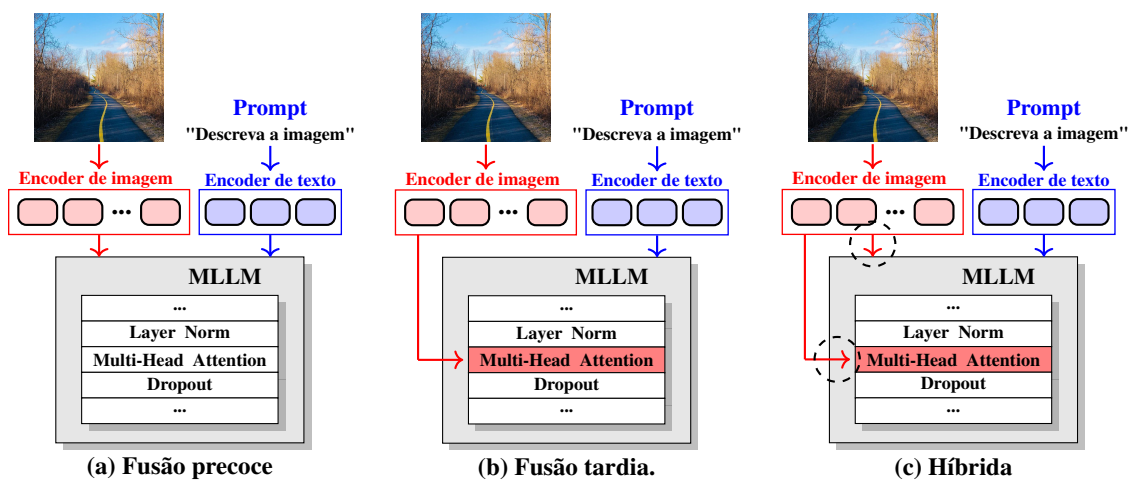

Dentro desse panorama, é possível classificar as estratégias de integração multimodal em termos de fusão precoce (também chamada de \textit{input-level fusion}) e fusão tardia (ou \textit{model-level fusion}). Na fusão precoce, as diferentes modalidades são alinhadas e combinadas em um espaço comum logo no início do processamento, de modo que o modelo aprenda representações conjuntas desde as camadas iniciais. Já na fusão tardia, cada modalidade é processada separadamente e apenas em estágios posteriores as representações são combinadas, tipicamente dentro do LLM. A Figura~\ref{fig:AlignmentSchemes} ilustra os dois tipos de integração, assim como um terceiro que agrega as duas estratégias, chamado de fusão híbrida. A escolha entre fusão precoce, tardia ou híbrida envolve importantes \textit{trade-offs} práticos em termos de custo computacional, reuso de modelos e capacidade de generalização. 

\textbf{Fusão precoce} oferece o maior grau de interação entre modalidades, pois o modelo aprende representações conjuntas desde as camadas iniciais. No entanto, essa abordagem exige grandes volumes de dados multimodais anotados e treinamento ponta a ponta, o que aumenta o custo e o tempo de ajuste fino. 

\textbf{Fusão tardia} é mais eficiente do ponto de vista de engenharia, pois permite reaproveitar \textit{encoders} e LLMs já pré-treinados. Ela requer apenas um módulo adicional para combinar as representações - frequentemente por meio de blocos de \textit{cross-attention} - mas pode gerar integrações mais superficiais, com menor sinergia entre modalidades.

\textbf{Fusão híbrida} busca equilibrar as limitações descritas anteriormente, combinando a integração inicial de \textit{embeddings} com mecanismos de fusão em camadas intermediárias. Isso tende a melhorar a flexibilidade e a coerência multimodal, embora introduza complexidade arquitetural e desafios de otimização. 

Um componente técnico central nas arquiteturas modernas de fusão tardia e híbrida é o mecanismo de atenção cruzada (\textit{cross-attention}). Diferentemente da auto-atenção, na qual cada token interage apenas com outros \textit{tokens} da mesma sequência, a atenção cruzada permite que \textit{tokens} textuais consultem representações visuais ou auditivas em cada etapa de decodificação. 

Em modelos como o \textit{Flamingo} \cite{alayrac2022flamingo}, blocos específicos de \textit{cross-attention} são intercalados nas camadas do LLM, funcionando como portas de comunicação entre modalidades. Esses blocos aprendem a pesar dinamicamente quais partes da imagem são mais relevantes para o texto sendo gerado, viabilizando raciocínio multimodal contextual. Essa abordagem preserva o núcleo linguístico do LLM intacto, mas adiciona uma camada de alinhamento adaptativo, na qual o modelo aprende não apenas a compreender a imagem, mas também a consultá-la seletivamente durante a geração textual (um aspecto essencial para tarefas como descrição visual, interpretação de gráficos e raciocínio \textit{visual-textual}).

\subsection{Modelos Emblemáticos e Arcabouços}
\label{subsec:modelos_emblematicos}

A trajetória recente dos MLLMs pode ser ilustrada por alguns modelos emblemáticos que marcaram diferentes fases de sua evolução. Entre eles, destaca-se o GPT-4V(\textit{ision}) \cite{openai2023gpt4v}, que estende o GPT-4 para processar imagens. Embora detalhes arquiteturais não sejam públicos, acredita-se que siga o paradigma de encapsulamento de modalidades, integrando um \textit{encoder} visual avançado ao núcleo de linguagem.  Sua contribuição reside na demonstração de que modelos comerciais de larga escala podem realizar raciocínio visual sofisticado, incluindo tarefas de OCR e interpretação contextual de imagens do mundo real.

O Gemini \cite{gemini2023report} representa outro marco, anunciado como a primeira família de modelos nativamente multimodais em sua concepção. Diferente de arquiteturas adaptadas, foi projetado desde o início para processar texto, imagem, áudio e vídeo de forma integrada. Sua arquitetura baseia-se em \textit{Transformers} modificados e treinados ponta a ponta em dados multimodais, permitindo uma interação mais profunda entre modalidades. A família de modelos Gemini inclui variantes otimizadas para diferentes contextos de uso, desde dispositivos móveis até grandes servidores de processamento.

O PaLM-E \cite{driess2023palmeembodiedmultimodallanguage}, desenvolvido pelo Google DeepMind, é um modelo multimodal voltado para robótica, que combina percepção visual e textual com capacidade de planejamento e tomada de decisão em ambientes físicos. Seu diferencial é demonstrar como os MLLMs podem ser incorporados em agentes encarnados, aproximando a inteligência artificial de aplicações práticas no mundo real.

\begin{figure}
    \centering
\centering
\includegraphics[width=0.9\linewidth]{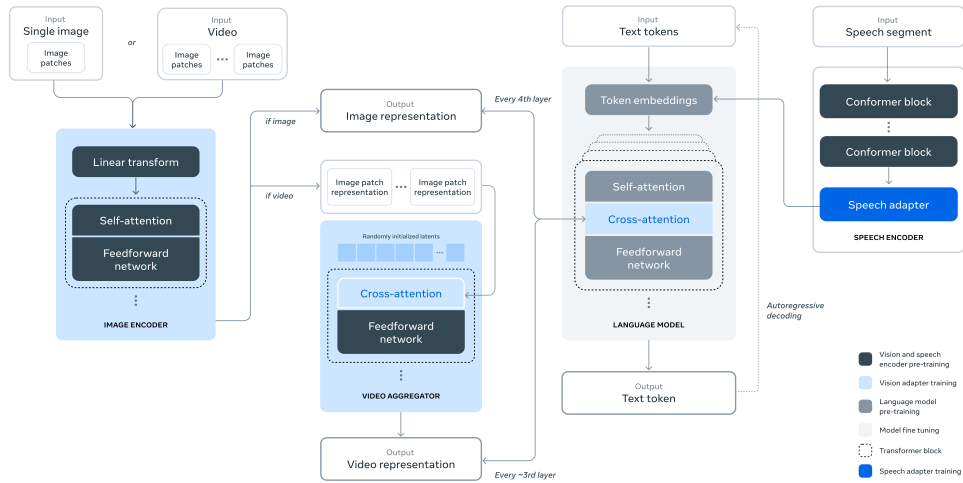}
\caption{Ilustração da abordagem de integração multimodal utilizada pelo Llama 3.2. Figura adaptada de \cite{grattafiori2024llama3herdmodels}.}
\label{fig:lhama}
\end{figure}

Outro modelo relevante é o Llama 3.2 \cite{grattafiori2024llama3herdmodels}, disponibilizado pela empresa Meta. Este modelo é um exemplo ilustrativo da estratégia de integração multimodal por fusão tardia, como mostrado na Figura \ref{fig:lhama}. Como se observa, além da modalidade textual, o modelo suporta também entradas visuais e sonoras.

Entre os modelos de código aberto, o LLaVA \cite{liu2023visual} tornou-se um dos mais influentes. Conectando o \textit{encoder} visual do CLIP a um LLM derivado do LLaMA  via projeção linear, demonstrou que a qualidade do alinhamento e o ajuste fino (\textit{fine-tuning}) em instruções visuais são mais determinantes que a escala da arquitetura. O BLIP-2 \cite{li2023blip} segue uma linha semelhante, mas introduz um módulo intermediário de projeção treinado para alinhar imagens a modelos de linguagem pré-existentes de forma mais eficiente. Essa abordagem reduziu significativamente o custo de treinamento multimodal, impulsionando a adoção em contextos de pesquisa e aplicações abertas.

\begin{figure}[ht]
\centering
\includegraphics[width=0.45\textwidth]{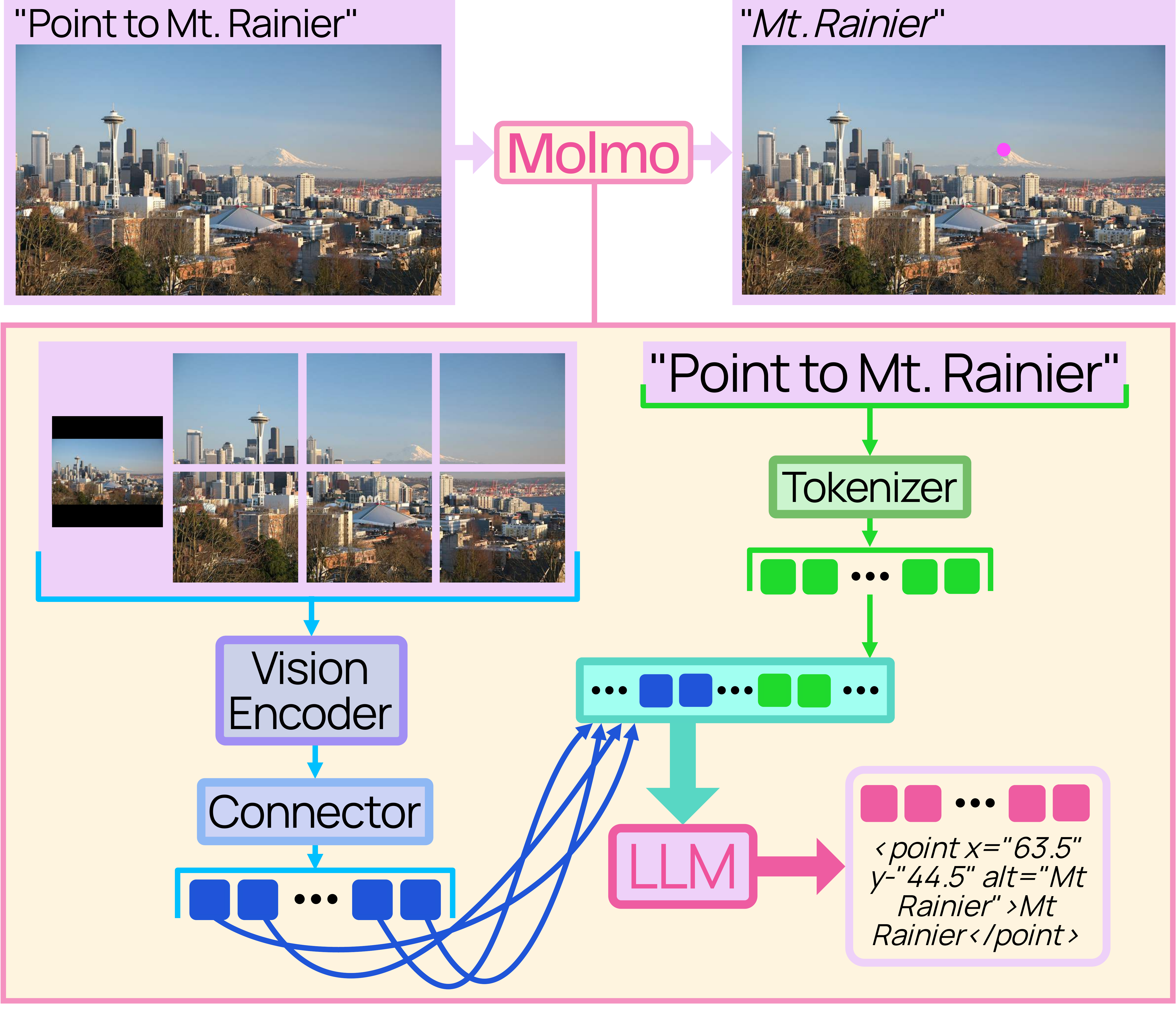}
\caption{O Molmo é um exemplo de modelo que adota a estratégia de fusão precoce para integração multimodal. Figura adaptada de \cite{deitke2024molmopixmoopenweights}.}
\label{fig:molmo}
\end{figure}

Outro modelo aberto é o Molmo \cite{deitke2024molmopixmoopenweights}. Conforme ilustrado na Figura \ref{fig:molmo}, o codificador de imagem utiliza um \textit{Transformer} de visão padrão, especificamente o CLIP. O termo \textit{conector} na figura refere-se a um \textit{projetor} que alinha os \textit{embeddings} da imagem com o modelo de linguagem, implementando uma fusão precoce entre as modalidades. Desenvolvido pela AllenAI, o Molmo foi lançado com pesos e dados abertos, buscando oferecer uma referência transparente e reprodutível para pesquisas em MLLMs.

\begin{figure}[ht]
\centering
\includegraphics[width=.8\textwidth]{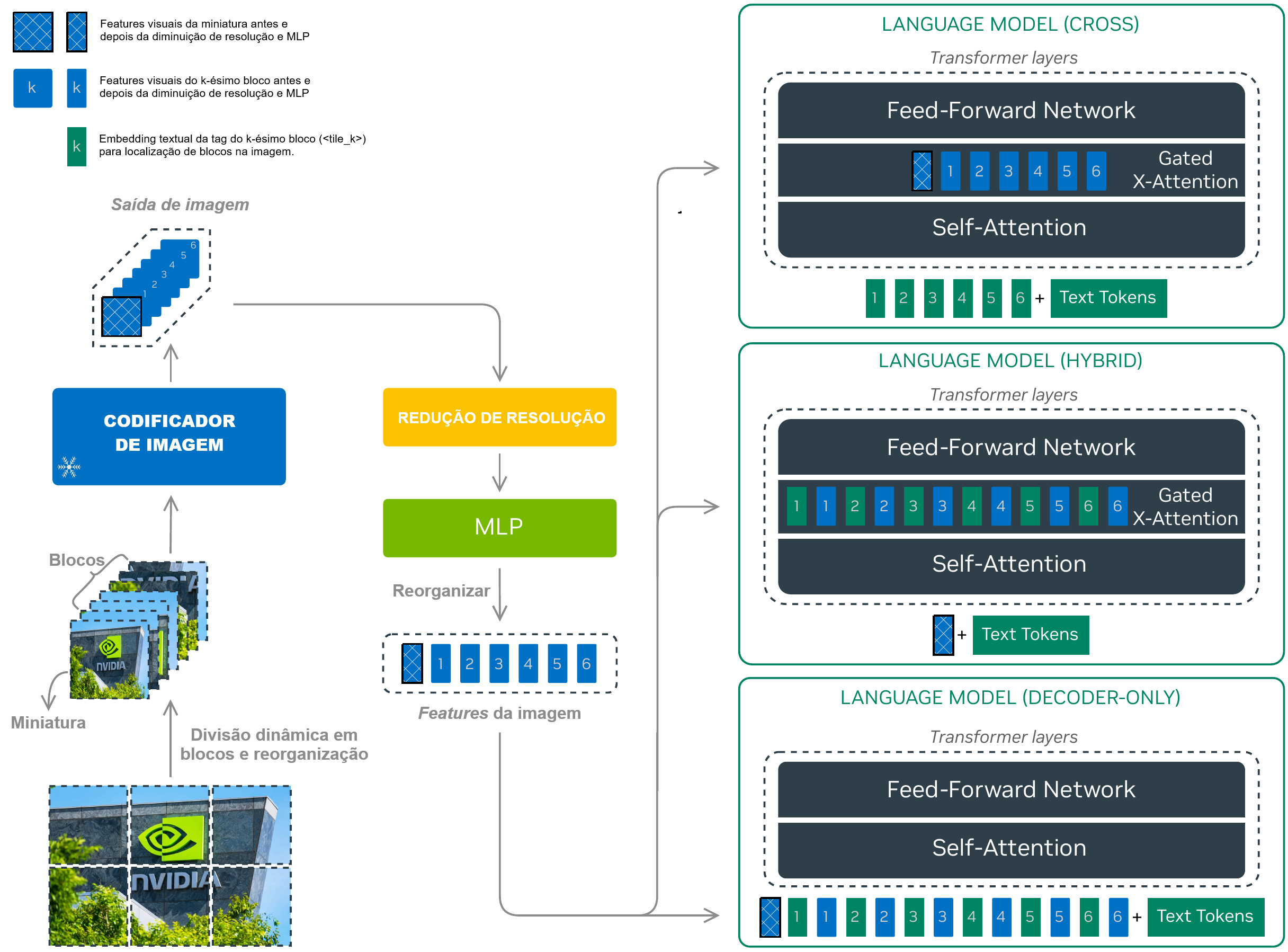}
\caption{O modelo NVLM-1.0 oferece três opções de arquiteturas: o NVLM-X baseado em atenção cruzada (superior), o NVLM-H híbrido (meio) e o NVLM-D somente decodificador (inferior). Figura adaptada de \cite{dai2024nvlmopenfrontierclassmultimodal}.}
\label{fig:nvidia}
\end{figure}

O modelo NVLM da NVIDIA \cite{dai2024nvlmopenfrontierclassmultimodal} é particularmente interessante porque, em vez de focar em uma única abordagem, explora ambas as estratégias de integração multimodal. Na Figura \ref{fig:nvidia}, a parte inferior nomeada \textit{Decoder-only} (NVLM-D) refere-se ao método de fusão precoce. A parte superior, nomeada \textit{Cross} (NVLM-X), refere-se ao método de fusão tardia. Além disso, os autores desenvolvem uma abordagem híbrida, nomeada \textit{Hybrid} (NVLM-H).

Os pesquisadores observam que a estratégia NVLM-X é mais eficiente, computacionalmente, para imagens de alta resolução. Já a estratégia NVLM-D apresenta melhor desempenho em tarefas relacionadas a OCR. Por fim, a estratégia NVLM-H combina os pontos fortes de ambos os métodos, NVLM-D e NVLM-X.

Por fim, ecossistemas como o Hugging Face \cite{wolf2020transformers} têm desempenhado papel crucial na disseminação dessas tecnologias. A biblioteca \texttt{transformers} fornece uma API unificada para carregar e utilizar modelos de linguagem e multimodais, além de \textit{pipelines} prontos para tarefas como geração de legendas de imagens, VQA e tradução multimodal. Esse ecossistema facilita o compartilhamento de modelos e dados, acelera a reprodutibilidade e democratiza o acesso aos MLLMs de última geração, reduzindo a barreira de entrada para instituições acadêmicas, pequenas empresas e  pesquisas independentes.

\section{Técnicas para Pipeline Multimodal }\label{sec:pipeline}

Esta seção apresenta os componentes fundamentais para a construção de \textit{pipelines} eficientes com modelos multimodais. O conteúdo é organizado de maneira progressiva, contemplando desde o tratamento inicial dos dados até ferramentas que permitem operacionalizar fluxos mais complexos. (i) Inicialmente, discute-se o pré-processamento multimodal, etapa essencial para garantir a qualidade das entradas e o alinhamento entre modalidades, permitindo que textos e imagens sejam representados em um espaço semântico compartilhado. (ii) Em seguida, são abordadas  algumas técnicas de engenharia de \textit{prompt}, incluindo estratégias como o \textit{chain-of-thought} multimodal (por exemplo, descrever o raciocínio sobre uma imagem de forma incremental) e o \textit{few-shot prompting} com pares imagem-texto para orientar o modelo em tarefas específicas. (iii) Por fim, a seção introduz arcabouços que facilitam a construção de \textit{pipelines} avançados, tais como LangChain e LangGraph. Tais arcabouços fornecem abstrações, componentes reutilizáveis, padronização de código e mecanismos de depuração e monitoramento, simplificando o desenvolvimento e a manutenção de sistemas multimodais.

\subsection{Pré-processamento de dados} \label{subsec:pipeline-preprocessing}

\begin{figure}[h]
    \centering
    \includegraphics[width=0.7\linewidth]{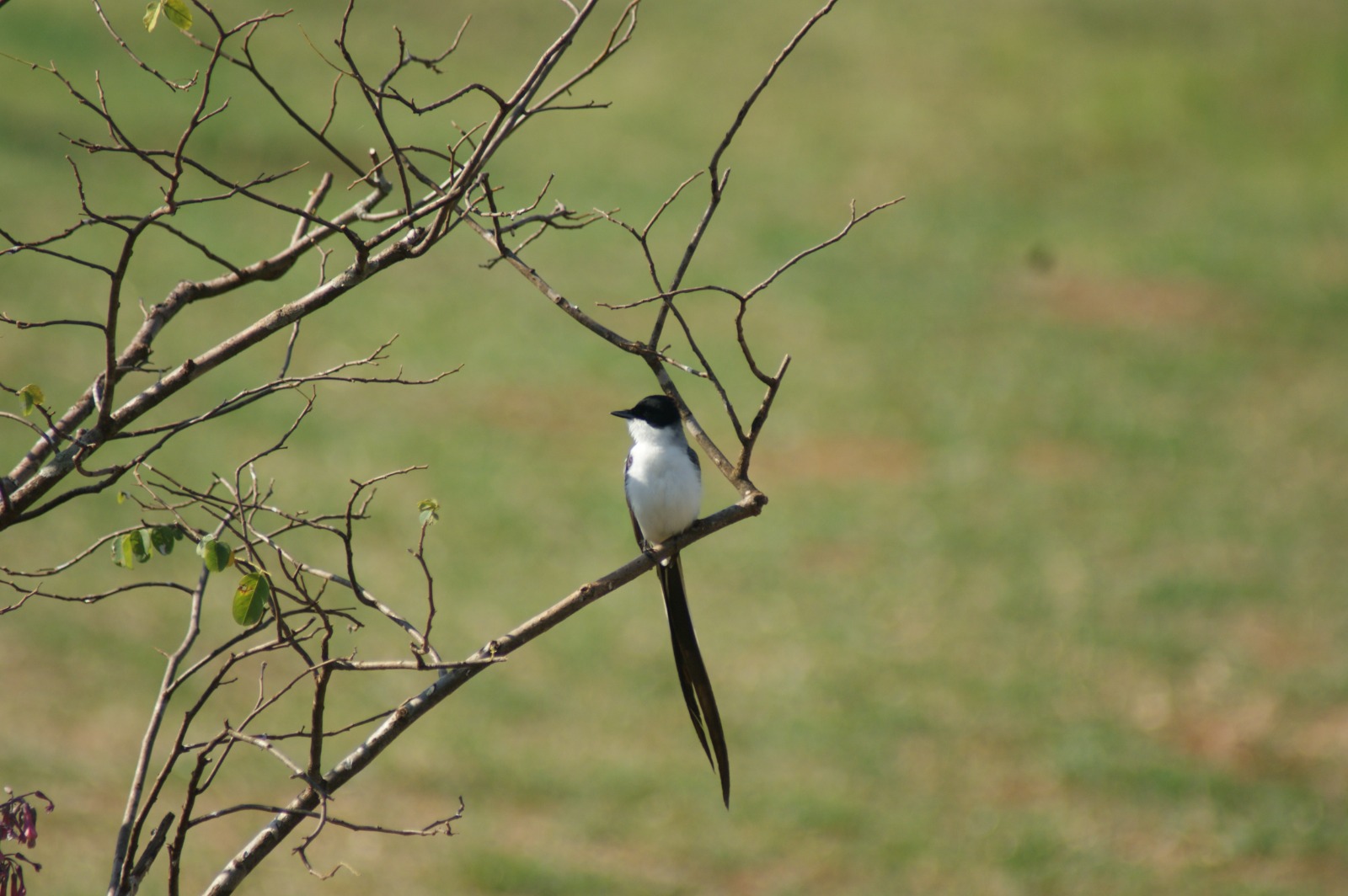}
  \caption{Fotografia da ave Tesourinha (nome científico: \textit{Tyrannus savana}). Fotografia cedida por Ricardo Gavira.}
  \label{fig:bird}
\end{figure}

O pré-processamento é uma etapa crucial no desenvolvimento de pipelines multimodais, pois é responsável por converter os dados brutos (imagens - ver exemplo na figura~\ref{fig:bird}, áudios, vídeos e textos) em representações vetoriais densas (\textit{embeddings}) compreensíveis pelos MLLMs. Em outras palavras, é nessa fase que se realiza  a ``tradução'' das modalidades para o espaço de linguagem do modelo.

Como discutido anteriormente, os MLLMs não operam diretamente sobre \textit{pixels}, sinais de áudio
ou \textit{tokens} crus, mas sobre vetores que representam semanticamente cada conteúdo. A etapa de pré-processamento, portanto, consiste em (i) normalizar e limpar os dados de entrada, (ii) extrair suas representações (via \textit{encoders} especializados), e (iii) projetar essas representações em um espaço semântico compatível com o LLM.

A implementação prática dessa etapa varia conforme a modalidade de entrada. Por exemplo, para mapear imagens e textos em um espaço multimodal compartilhado, há o modelo CLIP -- uma das abordagens mais utilizadas \cite{radford2021learning}. O Código \ref{lst:lst:preprocessing-clip} mostra como carregar o CLIP e gerar \textit{embeddings} da imagem apresentada na Figura \ref{fig:bird} e um texto para posterior integração no pipeline. 

\begin{center}
\begin{lstlisting}[language=Python]
from PIL import Image
import torch
import clip
from torchvision import transforms

# Carrega o modelo CLIP pré-treinado
device = "cuda" if torch.cuda.is_available() else "cpu"
model, preprocess = clip.load("ViT-B/32", device=device)

# Pré-processa uma imagem de entrada
image = preprocess(Image.open("bird.jpeg")).unsqueeze(0).to(device)

# Pré-processa o texto de entrada
text = clip.tokenize(["uma ave tesourinha em um galho de árvore"]).to(device)

# Extrai embeddings
with torch.no_grad():
    image_features = model.encode_image(image)
    text_features = model.encode_text(text)

# Normaliza e calcula similaridade
image_features /= image_features.norm(dim=-1, keepdim=True)
text_features  /= text_features.norm(dim=-1, keepdim=True)

similarity = (100.0 * image_features @ text_features.T).softmax(dim=-1)
print(f"Similaridade imagem-texto: {similarity.item():.4f}")
\end{lstlisting}
\captionof{listing}{Exemplo de implementação em Python utilizando o modelo CLIP para geração dos vetores $image\_features$ e $text\_features$ no mesmo espaço semântico (similaridade imagem-texto igual a 1.0).}
\label{lst:lst:preprocessing-clip}
\end{center}

O BLIP-2 \cite{li2023blip} é uma evolução do CLIP, projetado especificamente para conectar representações visuais a LLMs, usando um módulo intermediário chamado \textit{Q-Former}. Esse módulo aprende a projetar \textit{embeddings} de imagem para o espaço de linguagem. O Código \ref{lst:lst:preprocessing-blip} apresenta um código escrito em Python com o BLIP-2, utilizando a biblioteca \textit{Transformers} da \textit{Hugging Face}\footnote{\url{https://huggingface.co/}}. Note que neste exemplo, ao invés de calcular a similaridade, o modelo  gera a descrição da imagem (\textit{image captioning}).

\begin{center}
\begin{lstlisting}[language=Python]
from transformers import Blip2Processor, Blip2ForConditionalGeneration
from PIL import Image
import torch

device = "cuda" if torch.cuda.is_available() else "cpu"

# Carrega o modelo BLIP-2
processor = Blip2Processor.from_pretrained("Salesforce/blip2-flan-t5-xl")
model = Blip2ForConditionalGeneration.from_pretrained("Salesforce/blip2-flan-t5-xl", torch_dtype=torch.float16).to(device)

# Pré-processa a imagem
image = Image.open("exemplo_imagem.jpg")
inputs = processor(images=image, return_tensors="pt").to(device, torch.float16)

# Extrai embeddings e gera texto condicional
generated_ids = model.generate(**inputs, max_new_\textit{tokens}=20)
generated_text = processor.batch_decode(generated_ids, skip_special_\textit{tokens}=True)[0]
print("Descrição gerada:", generated_text)
\end{lstlisting}
\captionof{listing}{Exemplo de implementação em Python utilizando o modelo BLIP-2 para geração de descrição visual (\textit{captioning}), dada uma imagem de entrada.}
\label{lst:lst:preprocessing-blip}
\end{center}

Para dados de áudio, o CLAP (\textit{Contrastive Language–Audio Pretraining}) \cite{elizalde2023clap} desempenha papel semelhante ao CLIP, mapeando sons e descrições textuais para um espaço compartilhado, conforme mostrado no Código \ref{lst:lst:preprocessing-clap}.

\begin{center}
\begin{lstlisting}[language=Python]
from transformers import ClapProcessor, ClapModel
import torch
import torchaudio

# Carrega o modelo CLAP
processor = ClapProcessor.from_pretrained("laion/clap-htsat-unfused")
model = ClapModel.from_pretrained("laion/clap-htsat-unfused")

# Carrega um arquivo de áudio
waveform, sr = torchaudio.load("exemplo_audio.wav")

# Pré-processa o áudio e o texto
inputs = processor(audios=waveform, text=["som de pássaros cantando"], return_tensors="pt", padding=True)

# Extrai embeddings multimodais
with torch.no_grad():
    outputs = model(**inputs)
    audio_emb = outputs.audio_embeds
    text_emb = outputs.text_embeds
    similarity = outputs.logits_per_audio  # similaridade áudio-texto

print(f"Embedding de áudio: {audio_emb.shape}")
print(f"Embedding de texto: {text_emb.shape}")
print(f"Similaridade áudio-texto: {similarity.item():.4f}")
\end{lstlisting}
\captionof{listing}{Exemplo de implementação em Python utilizando o modelo CLAP para geração de \textit{embeddings} multimodais.}
\label{lst:lst:preprocessing-clap}
\end{center}

Além da geração de \textit{embeddings} multimodais, o pré-processamento pode incluir etapas que melhoram a qualidade e a interpretabilidade dos dados. Por exemplo, pode-se realizar a normalização, redimensionamento, segmentação de região de interesse (ROI do inglês \textit{Region of Interest}) de imagens; remoção de ruído e normalização de amplitude em áudios; tokenização, remoção de \textit{stopwords} e lematização em textos; geração de metadados contendo informações como formato, resolução, duração, ou anotações semânticas (e.g., entidades nomeadas), que podem auxiliar na recuperação e indexação de exemplos relevantes; e, ainda, o  aumento de dados (\textit{data augmentation}) multimodal, como variações visuais, ruído acústico ou reformulação de legendas, para aumentar a robustez do modelo. Tais práticas são especialmente importantes quando o objetivo é construir conjunto de dados multimodais personalizados.

Desta maneira, o pré-processamento é, portanto, a ponte entre os dados brutos e o raciocínio em múltiplas modalidades do MLLM. Ele transforma \textit{pixels}, sinais de áudio e \textit{tokens} em vetores comparáveis, alinhando múltiplas modalidades em um mesmo espaço de linguagem, permitindo que o modelo ``pense'' sobre imagens, sons e textos de forma integrada.

\subsection{Engenharia de \textit{Prompt}  para Tarefas Multimodais} \label{subsec:pipeline-prompteng}

Para melhorar os resultados dos MLLMs, um aspecto central é a elaboração de bons \textit{prompts}, processo conhecido como engenharia de \textit{prompt}. Mas o que caracteriza um ``bom'' \textit{prompt}? De modo geral, é um conjunto de comandos com instruções claras, específicas e contextualizadas, nas quais a tarefa é descrita de maneira objetiva e, quando necessário, acompanhada de exemplos. Além disso, o \textit{prompt} pode incluir instruções sobre o formato da saída (por exemplo, lista, tabela, JSON ou texto corrido) e orientações sobre estilo ou tom da resposta.
 A estruturação do \textit{prompt} desempenha um papel significativo, especialmente quando ele se torna longo. Nestes casos, recomenda-se organizar o conteúdo em blocos lógicos, como listas numeradas, seções com títulos, ou palavras-chave destacadas. \textit{Prompts} organizados facilitam o processamento interno do modelo e reduzem ambiguidades. Outra estratégia comumente utilizada é direcionar o modelo a assumir uma persona, isto é, uma configuração de comportamento ou papel atribuído ao modelo, influenciando seu estilo de resposta, tom, vocabulário e nível de detalhamento, como se estivesse ``interpretando'' uma identidade específica.

Com essas práticas, já é possível obter resultados relevantes por meio do \textit{zero-shot prompting}, que consiste em solicitar que o modelo realize uma tarefa sem fornecer exemplos prévios, apenas com instruções diretas. No entanto, em tarefas mais complexas, o \textit{zero-shot} pode não ser suficiente. Assim, utilizam-se técnicas  mais avançadas,  como o \textit{few-shot prompting}, \textit{Chain-of-Thought} (CoT), \textit{self-consistency}, ReAct (\textit{Reason} + \textit{Act}), e o \textit{prompting} multimodal guiado. 

O \textit{few-shot prompting} \cite{brown2020language} consiste em fornecer ao modelo alguns exemplos da tarefa desejada diretamente no \textit{prompt}, antes de solicitar a resposta final. Ao observar esses exemplos, o modelo identifica o padrão de entrada e saída, reproduzindo o estilo, abordagem e nível de detalhamento esperado. Essa técnica é especialmente útil quando a tarefa é complexa, ambígua ou envolve convenções específicas, por exemplo, formato de resumo, estilo narrativo, ou critérios específicos de classificação. Em modelos multimodais, o \textit{few-shot} pode incluir não apenas texto, mas imagens ou combinações imagem-texto, demonstrando como relacionar as modalidades. É importante escolher exemplos curtos, representativos e consistentes, pois exemplos ruins ou variados demais podem levar o modelo a reproduzir padrões incorretos.

Já o CoT \cite{wei2022chain} incentiva o modelo a explicitar o raciocínio passo a passo antes de apresentar a resposta final. Esse encadeamento de pensamento ajuda o modelo a organizar as informações de maneira sequencial e lógica, podendo reduzir erros e aumentando a precisão em tarefas que exigem raciocínio complexo, como problemas matemáticos, interpretação de gráficos, inferência causal e análise de imagens com múltiplos elementos. Em MLLMs, o CoT pode envolver a descrição progressiva da cena visual, destacando observações intermediárias antes de chegar à conclusão. A orientação explícita para ``pensar em voz alta'' tende a melhorar a qualidade da resposta, embora deva ser usada com cuidado em contextos sensíveis ou quando se deseja respostas curtas.

A técnica de \textit{self-consistency} \cite{wang2022self} parte da ideia de que o modelo pode gerar diferentes cadeias de raciocínio possíveis para uma mesma tarefa, especialmente quando o problema admite mais de um caminho lógico. Em vez de confiar em uma única resposta do CoT, o modelo é instruído a gerar múltiplas explicações ou raciocínios independentes, e posteriormente selecionar (ou combinar) o resultado mais recorrente ou coerente entre eles. Na prática, isso reduz o impacto de cadeias de raciocínio incorretas ou enviesadas que poderiam surgir em uma única tentativa. Apesar de aumentar a qualidade, ela implica maior custo computacional, já que requer múltiplas amostragens.

O ReAct \cite{yao2022react} é uma técnica que combina raciocínio verbal e ações iterativas para resolver problemas de forma interativa. O modelo não apenas descreve o raciocínio, mas também executa ações intermediárias, como consultar documentos, examinar imagens novamente, fazer buscas externas ou solicitar informações adicionais. Esse ciclo ``pensar - agir - pensar novamente'' torna o modelo mais adequado para funções de agente autônomo, pipelines multimodais e cenários em que é necessário integrar fontes externas de conhecimento. Em MLLMs, isso pode significar interpretar uma imagem, fazer hipóteses, verificar detalhes adicionais na própria imagem ou consultar metadados, e então ajustar a conclusão final. O ReAct reduz erros relacionados a suposições precipitadas, mas depende de um ambiente que permita operações de consulta ou APIs externas.

O \textit{prompting} multimodal guiado \cite{liu2023visual} refere-se a estratégia de orientar explicitamente como o modelo deve integrar e relacionar informações de diferentes modalidades. Por exemplo, pode-se instruir o modelo da seguinte forma: ``descreva o que está na imagem e explique como isso se relaciona com o texto fornecido''. Em MLLMs, simplesmente fornecer texto e imagem não garante que o modelo interpretará suas relações de forma alinhada ao objetivo da tarefa. Por isso, o \textit{prompt} pode orientar etapas, como observar a imagem primeiro, identificar elementos-chave, relacionar esses elementos ao texto fornecido, e apenas então formular a resposta final. Essa técnica melhora a coerência entre modalidades e reduz interpretações equivocadas ou superficialmente descritivas. Ela é particularmente útil em tarefas de análise de documentos visuais, gráficos científicos, \textit{captioning}, e explicação de conteúdo visual.

\subsection{Arcabouços para facilitar a construção de \textit{pipelines} avançados}
\label{subsec:pipeline-frameworks}

\begin{figure}[ht]
  \centering
    \includegraphics[width=.5\linewidth]{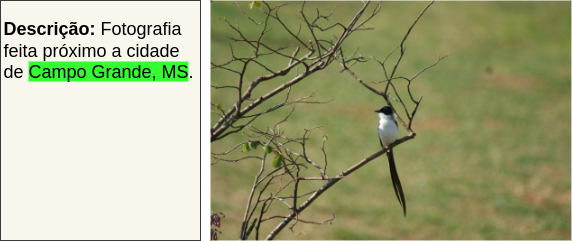}
  \caption{Exemplo de entrada para o MLLM, cuja imagem é a mesma presente na Figura \ref{fig:bird} e a entidade nomeada referente ao local está em destaque no texto.}
  \label{fig:example-input}
\end{figure}

A depender da complexidade da tarefa, os MLLMs, por si só, podem não ser suficientes, tornando necessária a criação de um \textit{pipeline} que combine esses modelos com outras ferramentas para alcançar o objetivo desejado de forma satisfatória. Por exemplo, considere a plataforma WikiAves\footnote{https://www.wikiaves.com.br/. Acessado em 9 de outubro de 2025.}, uma conhecida comunidade \textit{online} de observadores de aves que mantém uma extensa base de dados sobre aves do Brasil. Suponha que essa plataforma deseje adicionar uma funcionalidade de busca por imagens, permitindo que os usuários façam \textit{upload} de uma foto acompanhada de um breve texto com informações adicionais, como a localidade onde o registro foi feito, conforme ilustrado na Figura \ref{fig:example-input}. Dessa forma, mesmo usuários iniciantes poderiam aprender mais sobre as aves, ainda que não conheçam seus nomes, que atualmente constituem o principal critério de busca na plataforma.

Para isso, a WikiAves pretende utilizar um MLLM capaz de processar as informações fornecidas pelo usuário (isto é, a foto e o breve texto) para identificar a espécie da ave e, em seguida, buscar em sua base de dados o registro correspondente, redirecionando o usuário para a página que contém todas as informações sobre ela. Considerando o exemplo apresentado na Figura \ref{fig:example-input}, realizou-se um teste utilizando o modelo GPT-4o-mini-2024-07-18, da OpenAI (ou apenas, GPT-4o-mini, por simplicidade), por meio de uma requisição à API\footnote{https://api.openai.com/v1/chat/completions. Utilizado em 9 de outubro de 2025.} do modelo, representada no Código \autoref{lst:openai-api-gpt4o-mini}.

\begin{center}
\begin{lstlisting}[language=json]
{
  "model": "gpt-4o-mini",
  "messages": [
    {
      "role": "system",
      "content": "Você é um assistente especializado em análise visual. Você recebe como entrada uma imagem de uma ave e um breve texto com informações sobre a foto, e deve identificar qual ave se trata e responder o seu nome popular e científico. Responda em português."
    },
    {
      "role": "user",
      "content": [
        {
          "type": "text",
          "text": "Fotografia feita próximo a cidade de Campo Grande, MS."
        },
        {
          "type": "image_url",
          "image_url": {
            "url": "<URL>"
          }
        }
      ]
    }
  ],
  "max_\textit{tokens}": 500
}
\end{lstlisting}
\captionof{listing}{Exemplo de \textit{payload} JSON para requisição à OpenAI API, no qual é preciso substituir <URL> por uma URL válida.}
\label{lst:openai-api-gpt4o-mini}
\end{center}

\begin{figure}[ht]
  \centering
    \includegraphics[width=\linewidth]{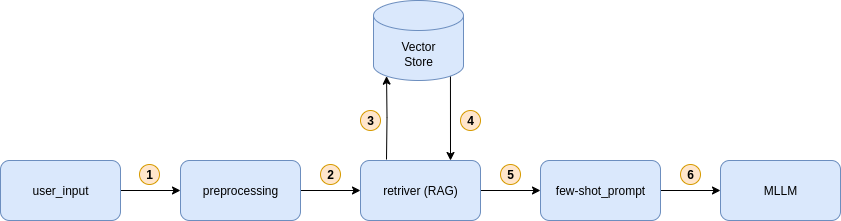}
  \caption{Exemplo de \textit{pipeline} para possibilitar a busca por imagens.}
  \label{fig:example-mllm-pipeline}
\end{figure}

Obteve-se a seguinte resposta do modelo: ``A ave na fotografia é o \textbf{Bem-te-vi}. Seu nome científico é \textbf{Pitangus sulphuratus}'', informações estas claramente erradas. Os motivos que levaram o modelo a cometer esse erro podem ser diversos, incluindo a possível falta de conhecimento sobre a fauna brasileira durante seu pré-treinamento. Para superar essa limitação, a plataforma WikiAves precisaria implementar um \textit{pipeline} mais sofisticado para aprimorar o desempenho do modelo. A Figura \ref{fig:example-mllm-pipeline} apresenta um possível \textit{pipeline} para mitigar esse problema.

Como pode-se observar, após obter a entrada fornecida pelo usuário, o primeiro passo do \textit{pipeline} é o pré-processamento, que, conforme discutido na Seção \ref{subsec:pipeline-preprocessing}, além de gerar os \textit{embeddings}, pode também ser utilizado para melhorar a qualidade da imagem, identificar e segmentar a região de interesse (isto é, a porção da imagem em que se encontra a ave) e extrair as entidades nomeadas presentes no texto, neste caso, o local onde a fotografia foi capturada (a cidade de Campo Grande, MS). Com essas informações em mãos, o próximo passo do \textit{pipeline} consiste em recuperar, a partir da base de dados da WikiAves, previamente vetorizada, ou seja, a \textit{vector store}, registros de aves semelhantes que podem ocorrer na região informada, utilizando algum método de RAG (\textit{Retrieve Augmented Generation}) \cite{lewis2020retrieval}. A partir dos resultados obtidos nessa etapa, constrói-se o \textit{few-shot prompting}, como descrito na Seção \ref{subsec:pipeline-prompteng}, no qual são inseridos no \textit{prompt} os exemplos de aves semelhantes recuperados via RAG. Dessa forma, o modelo pode aprender em tempo de execução com os exemplos fornecidos no contexto, para elaborar uma resposta mais assertiva.

Pode parecer um processo bem mais complexo do que simplesmente realizar uma solicitação à API do modelo, mas atualmente existem diversos arcabouços que facilitam a construção de \textit{pipelines} como esse, tais como LangChain e LangGraph. Esses arcabouços fornecem várias ferramentas já implementadas, o que torna seu uso bastante simples, além de contribuírem para a padronização do código e oferecerem recursos para depuração de eventos.

No restante desta seção, serão fornecidos detalhes da utilização do LangChain e do LangGraph, arcabouços amplamente adotados pela indústria e pela academia, para a construção de \textit{pipelines} com MLLMs.

\subsubsection{LangChain} \label{subsubsec:langchain}
Com o LangChain \cite{langchain2025}, é possível modelar um \textit{pipeline} multimodal como uma sequência de componentes encadeados (\textit{chains}), em que cada etapa realiza uma tarefa específica e passa os resultados para a próxima.

Cada etapa pode ser implementada como uma função encapsulada em um \emph{RunnableLambda} ou \emph{LLMChain}, compondo uma cadeia sequencial com o \emph{SequentialChain}. O Apêndice \ref{appendix:langchain} mostra  um exemplo em alto nível, em Python, de como implementar o \textit{pipeline} utilizando o LangChain.

Durante a execução, o estado do \textit{pipeline} é atualizado a cada etapa, e seus campos (como roi, rag\_docs e \textit{prompt}) são propagados entre as funções. Essa abordagem torna o fluxo transparente e modular, permitindo substituir etapas individuais sem afetar as demais, incluir ramos de decisão e verificação de erros, integrar modelos e ferramentas externas (como \textit{embedders} multimodais, APIs e bancos vetoriais).
Além disso, o LangChain possui integrações nativas com múltiplos tipos de modelos LLMs, MLLMs, \textit{embeddings}, \textit{retrievers}, ferramentas de visão computacional, e mais, o que simplifica o desenvolvimento de pipelines híbridos \cite{langchain2025}.

A combinação de LangChain com ferramentas como LangSmith \cite{langsmith2025} facilita a instrumentação, rastreamento e depuração de cada etapa do \textit{pipeline}, sendo possível visualizar o fluxo completo de execução, incluindo as entradas e saídas de cada \textit{chain}; medir latência, custos de chamadas e métricas de desempenho; identificar gargalos, falhas e respostas inesperadas em tempo real.

Modelar o \textit{pipeline} com LangChain traz vantagens práticas e conceituais, tais como a \textbf{modularidade}, sendo que cada componente é independente, facilitando manutenção e testes; o \textbf{reuso}, onde as \textit{chains} podem ser combinadas ou reutilizadas em outros fluxos; a escalabilidade, já que permite adicionar novas etapas à \textit{chain}; e, por fim, a \textbf{integração}, tendo interoperabilidade direta com LangSmith, LangServe e LangStudio. Contudo, o LangChain apresenta algumas limitações em cenários que exigem controle mais sofisticado de fluxo e estado.

Por exemplo, cada \textit{chain} é essencialmente uma função isolada, e o estado global do \textit{pipeline} precisa ser propagado manualmente entre as etapas, o que aumenta a complexidade e o risco de inconsistências, especialmente em \textit{pipelines} multimodais nos quais \textit{embeddings} de diferentes tipos de dados (como imagem, texto e áudio) 
precisam ser sincronizados. Além disso, fluxos condicionais e ramificações lógicas não são tratadas de forma nativa. Implementações que envolvem verificações ou caminhos alternativos, como o caso em que nenhuma ave é detectada na imagem, acabam exigindo código imperativo fora da estrutura de \textit{chains}, reduzindo a clareza e modularidade do projeto. Outra limitação é a ausência de uma representação gráfica explícita do \textit{pipeline}, o que dificulta a visualização e a depuração em fluxos complexos. 

\subsubsection{LangGraph} \label{subsubsec:langgraph}

Em contrapartida às limitações do LangChain, o LangGraph \cite{langgraph2025}  permite modelar o \textit{pipeline} como um grafo de estados, no qual cada nó representa uma etapa e cada aresta define as condições de transição. Isso simplifica a manutenção, facilita a instrumentação e oferece suporte nativo a laços (\textit{loops}), subgrafos e fluxos condicionais.
Por esse motivo, o LangGraph se mostra mais adequado para \textit{pipelines} multimodais com múltiplas dependências ou ramos de execução, proporcionando maior clareza estrutural, controle de estado compartilhado e escala organizacional do que o modelo sequencial tradicional do LangChain.
Com o LangGraph \cite{langgraph2025}, o \textit{pipeline} é modelado como um grafo de estados (ou nós), em que cada nó representa uma etapa específica do fluxo de processamento. Caso ocorra uma condição de erro em qualquer etapa, por exemplo, nenhuma ave detectada na imagem durante o pré-processamento, é possível definir um fluxo condicional que encerra a execução do \textit{pipeline} de forma controlada.

Cada nó pode ler e/ou atualizar o estado compartilhado (\textit{State}) do grafo. O LangGraph oferece recursos para definir arestas condicionais, laços, subgrafos, entre outros mecanismos que tornam o \textit{pipeline} mais flexível e modular \cite{langgraph2025}.

Considerando o exemplo de \textit{pipeline} ilustrado na Figura \ref{fig:example-mllm-pipeline}, são apresentados no Apêndice \ref{appendix:langgraph} trechos de código com uma implementação em alto nível, escrita em Python, utilizando o arcabouço LangGraph.

Uma vez configurado o grafo, o LangGraph executa o \textit{pipeline} de forma determinística e observável, permitindo inspecionar o fluxo de dados e as transições entre nós. Durante a execução, o estado é propagado entre os nós, de modo que cada etapa pode consumir os resultados anteriores e produzir novas saídas, enriquecendo progressivamente o contexto compartilhado.

Essa representação em grafo oferece modularidade, facilitando a substituição, atualização ou reuso de componentes individuais, por exemplo, trocar o mecanismo de recuperação. Também simplifica a instrumentação e o monitoramento, permitindo identificar gargalos de execução, métricas de desempenho e resultados intermediários. Ferramentas como o LangSmith são amplamente utilizadas nesse contexto, pois oferecem recursos avançados de rastreamento, depuração e visualização interativa dos fluxos de execução, facilitando o entendimento de como os dados percorrem o grafo e como cada nó contribui para o resultado final.

Por fim, o LangGraph integra-se naturalmente ao ecossistema LangChain, tornando possível combinar agentes, memória e ferramentas adicionais dentro do mesmo fluxo. Isso torna o arcabouço especialmente adequado para \textit{pipelines} multimodais complexos, em que há múltiplas fontes de dados e etapas de raciocínio encadeadas.

\section{Aplicações Práticas e Casos de Uso}\label{secAppPraticas}

Os MLLMs representam um dos avanços mais expressivos na confluência entre processamento de linguagem natural, visão computacional e aprendizado profundo. Após a consolidação teórica e o amadurecimento das arquiteturas que integram texto, imagem, áudio e vídeo, torna-se essencial compreender como esses modelos se traduzem em aplicações concretas. Esta seção discute exemplos reais de uso dos MLLMs em diferentes contextos, evidenciando sua versatilidade e potencial para transformar a interpretação e interação desses modelos com dados multimodais.

\subsection{Análise em Conjunto de Dados Públicos} 

A disponibilidade de conjunto de dados públicos contribui para o avanço tecnológico, ao habilitar a realização de testes de hipóteses sobre integração semântica entre múltiplas modalidades no contexto dos MLLMs. Essa disponibilidade possibilita o desenvolvimento de modelos em larga escala com maior capacidade de generalização e desempenho cada vez mais robusto. No entanto, a qualidade, o tamanho e a diversidade dos conjunto de dados exigem atenção especial, uma vez que esses fatores afetam de forma decisiva a habilidade do modelo em compreender os dados e produzir representações consistentes.

Conjuntos como \textit{Visual Genome} estenderam a noção de  \textit{dataset} mutimodal ao incluir relações semânticas entre objetos, atributos e regiões específicas da imagem. Esse avanço permite aos MLLMs uma evolução da compreensão de contextos visuais de forma relacional e não apenas descritiva, podendo ser denominadas de \textit{visual grounding} e \textit{scene graph generation} \cite{krishna2017}.

No campo das emoções, conjunto de dados como o \textit{PerceptSent} \cite{PerceptSent} se destacam por incorporarem anotações afetivas em imagens permitindo análises mais profundas além do conteúdo objetivo. O conjunto de dados explora como as pessoas atribuem emoções a diferentes estímulos visuais consolidando a saída com informações quantitativas, pontuação do sentimento atribuído e perfil do avaliador, e até qualitativa, breves comentários sobre a explicação do sentimento ou motivações para a pontuação do sentimento. Essa abordagem híbrida amplia o potencial dos MLLMs em tarefas como análise de humor em redes sociais e psicologia computacional.

No entanto, é importante notar que embora a disponibilidade desses conjunto de dados tenha impulsionado diversos ramos de pesquisas, suas utilizações requerem análise crítica. Questões como representatividade de gênero e cultura, curadoria automatizada e licenciamento ético têm sido foco de debate  \cite{birhane2021}. A adoção de dados públicos, ou não, deve ser acompanhada de práticas de avaliação e filtragem, garantindo que o aprendizado multimodal não reproduza vieses ou distorções perceptuais.

\subsection{Geração de descrições de imagens} 
A geração de descrições a partir de imagens, conhecida pelo termo \textit{image captioning}, é uma das aplicações de MLLMs que vai além da identificação básica de objetos em uma imagem mas envolve a interpretação do contexto, inferir relações e traduzir percepções visuais em linguagem natural coerente e contextualizada.

Historicamente, os primeiros avanços para a geração de descrições de imagens ocorreram com o uso de arquiteturas \textit{enconder-decoder} baseadas em aprendizado profundo (\textit{deep learning}) nas quais a rede neural convolucional extraía características visuais e uma rede neural recorrente, ou LSTM, era responsável por gerar a sequência textual \cite{vinyals2015show,hossain2019survey}. Entretanto arquiteturas como essas apresentavam limitações em capturar nuances semânticas mais complexas.

Com a transição para o uso de arquiteturas baseadas em \textit{Transformers} o resultado na qualidade das descrições foi aprimorado, entregando resultados mais consistentes e dinâmicos. Modelos que incorporam mecanismos de atenção cruzada, permitem correlacionar regiões específicas da imagem com palavras ou conceitos relevantes do texto, por exemplo, o modelo BLIP-2 \cite{li2023blip}. Com o surgimento dos MLLMs, modelos avançados, como o GPT-4V, possibilitam o uso de técnicas de \textit{captioning} contextual, nas quais é possível gerar diversos tipos de versões de \textit{outputs} do modelo, ou seja, descrições mais humorísticas ou mais técnicas sobre a mesma imagem \cite{openai2023gpt4}.

Como exemplo de aplicação prática, destacam-se modelagens como o Flamingo, desenvolvido pela DeepMind, que introduz a capacidade \textit{few-shot} na geração multimodal \cite{alayrac2022flamingo}. Diferentemente de modelos que demandam retreinamento, o Flamingo é capaz de adaptar-se rapidamente a novos domínios a partir de um número reduzido de referências. Esse recurso viabiliza aplicações práticas em contextos especializados, como descrição automatizada de imagens médicas e anotação assistida em comércio eletrônico, onde o estilo e vocabulário de descrição variam conforme o domínio.

Pesquisas recentes têm explorado o uso de \textit{image captioning} como ferramenta auxiliar para aprimorar os modelos multimodais. Propostas como o \textit{Vision-Language Reinforcement Model} utilizam uma abordagem baseada em reforço para refinar a qualidade das legendas, atribuindo recompensas a descrições semanticamente ricas e penalizando redundâncias 
\cite{dzabraev2024vlrm}. Em demostrações práticas, o modelo produziu legendas capazes de expressar elementos subjetivos e emocionais, como tons de humor, clima ou expressões faciais, um passo relevante para tecnologias de acessibilidade digital, como o aplicativo \textit{Be My Eyes}, que usa \textit{captioning multimodal} para descrever imagens a pessoas com deficiência visual \cite{bemyeyes2023}.

Dessa forma, a tarefa de geração de descrições de imagens não apenas constitui um campo consolidado de aplicações práticas, mas também serve como base metodológica para as demais tarefas multimodais, como análise de sentimentos visuais ou até mesmo VQA, tarefa na qual a etapa de descrição contextual frequentemente antecede a inferência.

\subsection{Tarefa de VQA} \label{subsec:preprocessing}
VQA sintetiza o maior objetivo do uso de MLLMs: a compreensão da conexão entre a interpretabilidade da imagem e a geração do questionamento realizado sobre ela, formulado em linguagem natural, integrando o raciocínio visual e textual.

Por exemplo, diante de uma imagem urbana, perguntas como ``Quantos carros estão estacionados?'' ou ``Qual é o estilo arquitetônico predominante?'' envolvem tanto percepção visual, na identificação e contagem de elementos, quanto na interpretação semântica para o reconhecimento cultural.

\begin{figure}[!htb]
  \centering
    \includegraphics[width=.5\linewidth]{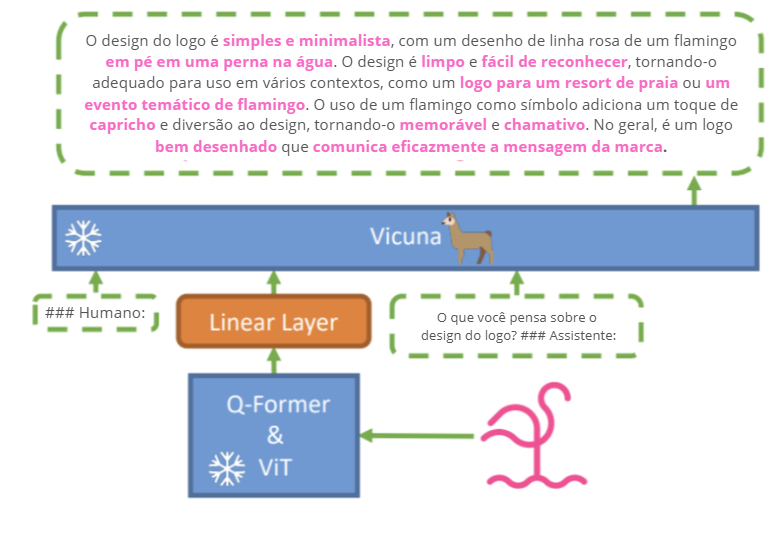}
  \caption{Exemplo de Arquitetura geral composta por \textit{vision encoder} do modelo MiniGPT-4  pré treinado conectado a um LLM, estrutura que possibilita o raciocínio multimodal e o \textit{captioning} contextual. composta por \textit{vision encoder} de \cite{zhu2023minigpt4}.}
  \label{fig:mini_gpt4}
\end{figure}

Esses sistemas baseiam-se em \textit{pipelines} compostos por módulos de percepção visual, codificação linguística e fusão semântica, como é possível observar na Figura \ref{fig:mini_gpt4}. Modelos como o LLaVA  \cite{liu2023visual} e MiniGPT-4 \cite{zhu2023minigpt4} exemplificam essa abordagem. Eles utilizam \textit{embeddings} compartilhados que projetam textos e imagem em um espeço vetorial unificado, no qual o raciocínio cruzado ocorre de forma natural. 

As aplicações práticas para tarefas de perguntas e respostas são vastas. Na área da saúde, por exemplo, modelos multimodais podem responder perguntas sobre exames de imagem, auxiliando médicos em diagnósticos preliminares \cite{Gong2025_MedBLIP}.
Na área de segurança e monitoramento, modelos multimodais podem interpretar cenas em tempo real e identificar comportamentos anômalos~\cite{Li2024_MultiImageVQA}. Já na educação interativa, sistemas de VQA permitem um aprendizado visual dinâmico, no qual estudantes exploram imagens científicas ou históricas por meio de perguntas em linguagem natural \cite{Pal2025_MultilingualVQA}.

Apesar dos avanços no domínio de perguntas e respostas multimodais, ainda persistem desafios relacionados ao viés de raciocínio e à dependência contextual. No que diz respeito ao viés de raciocínio, o modelo tende a responder com base em correlações estatísticas, em vez de realizar inferências visuais efetivas. Já a dependência contextual refere-se ao fato de que pequenas variações linguísticas na formulação da pergunta podem alterar significativamente a resposta produzida pelo MLLM. Para mitigar essas limitações, estudos recentes têm combinado modelos de raciocínio simbólico com arquiteturas multimodais, buscando integrar a robustez estatística dos MLLMs com capacidades de interpretação lógica.  Essa combinação busca conferir ao modelo maior precisão e maior interpretabilidade~\cite{Goyal2017_MakingVmatter}.

\subsection{Aplicações em Domínios Específicos: Análise de Sentimentos em Imagens (Arcabouço \textit{MLLMsent}) } \label{subsec:preprocessing}

Além das tarefas clássicas, os MLLMs têm demonstrado desempenho promissor em domínios especializados, nos quais a integração entre diferentes modalidades amplia a capacidade de análise de fenômenos complexos. Um desses domínios é a análise de sentimentos em imagens, campo em expansão no contexto da mineração \textit{multimodal} de dados. Nessa abordagem, elementos visuais, como expressões faciais, composição de cores e contexto, são combinados a informações textuais, como legendas ou comentários, para inferir o tom emocional de uma imagem. Estudos indicam, por exemplo, que a correlação entre a paleta cromática de uma imagem e o tipo de emoção expressa no texto associado pode revelar padrões culturais e psicológicos relevantes \cite{borth}.

Diante desse cenário, torna-se relevante compreender como os MLLMs têm aprimorado essa tarefa, substituindo abordagens puramente visuais por métodos capazes de integrar visão e linguagem de forma mais interpretável e contextualizada. A seguir, apresenta-se uma visão geral sobre os principais avanços na análise de sentimentos em imagens e o papel dos MLLMs nesse processo.

Diferente da classificação de objetos,  a \textbf{análise de sentimentos em imagens}  exige que o modelo compreenda o contexto, a emoção e as nuances transmitidas visualmente~\cite{outdoorsent2020,PerceptSent}. Trabalhos recentes, como o arcabouço \textit{MLLMsent} proposto por \cite{da2025multimodal}, investigam arquiteturas para solucionar este problema, focando em uma questão central: é mais eficaz classificar o sentimento diretamente da imagem ou a partir de uma descrição textual gerada pelo modelo?

\begin{figure}[!htb]
\centering

\begin{tikzpicture}[scale=0.98, every node/.style={scale=0.98}]
\tikzset{blockNA/.style={draw, rectangle, text centered, drop shadow, fill=red!20!white, text width=1.5cm, minimum height=0.8cm, inner sep=2pt}}

\tikzset{blockNB/.style={draw, rectangle, text centered, drop shadow, fill=blue!20!white, text width=1.5cm, minimum height=0.8cm, inner sep=2pt}}

\tikzset{blockMA/.style={draw, rectangle, text centered, drop shadow, fill=green!20!white, text width=5.5cm, minimum height=0.8cm, align=center, inner sep=4pt}}

\tikzset{blockM/.style={draw, rectangle, text centered, drop shadow, fill=white, text width=5.5cm, minimum height=0.8cm, align=center, inner sep=4pt}}

\tikzset{blockL/.style={draw, rectangle, text centered, drop shadow, fill=white, text width=2.8cm, minimum height=0.8cm, align=center, inner sep=4pt}}

\path[->](2.0,10.6) node[] (s1) {\scriptsize \textbf{Prompt}};
\draw(2.0,10.0) node[inner sep=0pt] (eval) {
\includegraphics[scale=0.4]{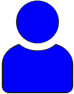}
};
\draw(2.8,10.0) node[inner sep=0pt] (prompt) {
\includegraphics[scale=0.13]{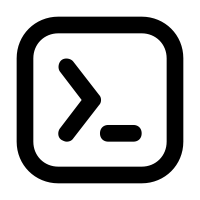}
};

\path[->](4.5,10.87) node[] (s1) {\scriptsize \textbf{Imagem de entrada}};
\draw(4.5,10.05) node[inner sep=0pt] (image) {
\includegraphics[width=2.0cm,height=1.3cm]{figs/bird.jpeg}
};

\path[->](3.2,8.6) node[blockMA] (llm) {
\scriptsize \textbf{Grande Modelo de\\[-0.5em] Linguagem Multimodal (MLLM)} 
};

\draw[->] (eval.south) -- ($(eval.south)!(llm.north)!(eval.south)$);

\draw[->] (image.south) -- ($(image.south)!(llm.north)!(image.south)$);

\path[->](3.2,7.2) node[blockL] (ic) {
\scriptsize\textbf{Classificação Direta\\[-0.5em] de Imagens} 
};

\path[->](3.2,5.85) node[blockM] (target) {
\scriptsize \textbf{Análise de sentimentos\\[-0.5em] (polaridades)} 
};

\draw(5.4,5.84) node[inner sep=0pt] (image) {
\includegraphics[width=1.0cm,height=0.9cm]{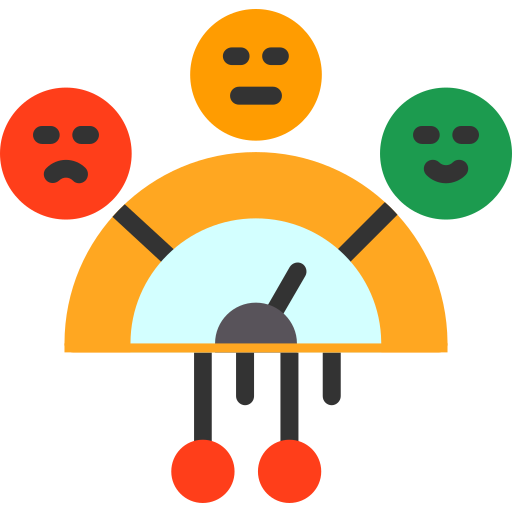}
};

\draw[->] (llm) -- ($(llm.south)!(ic.north)!(llm.south)$);

\draw[->] (ic) -- ($(ic.south)!(target.north)!(ic.south)$);

   \node at (7.8,8.5) 
    [align=center, text width=2.5cm, text height=0.2cm, fill=green!20!white, dotted, inner sep=7pt, thick] (versopt) 
    {
    \hspace{-20pt}
      \begin{minipage}[t][1.5cm][t]{1.1\textwidth}
        \centering
        \begin{itemize}
         \setlength{\itemsep}{3pt}
         \setlength{\labelsep}{4pt} 
         \setlength{\leftmargin}{-20pt} 
          \scriptsize
          \item {\VLMos} \vspace{-7pt}
          \item {\VLMoai} \vspace{-7pt}
          \item {\VLMds} \vspace{-7pt}
          \item Gemini \vspace{-7pt}
          \item $\LARGE \dots$
        \end{itemize}
      \end{minipage}
    };

\path[->](3.0,4.87) node[] (s1) {\small \textbf{(a)} };

\end{tikzpicture} \vspace{20pt}

\begin{tikzpicture}[scale=0.9, every node/.style={scale=0.98}]
\tikzset{blockNA/.style={draw, rectangle, text centered, drop shadow, fill=red!20!white, text width=1.5cm, minimum height=0.8cm, inner sep=2pt}}

\tikzset{blockNB/.style={draw, rectangle, text centered, drop shadow, fill=blue!20!white, text width=1.5cm, minimum height=0.8cm, inner sep=2pt}}

\tikzset{blockMA/.style={draw, rectangle, text centered, drop shadow, fill=green!20!white, text width=5.5cm, minimum height=0.8cm, align=center, inner sep=4pt}}

\tikzset{blockM/.style={draw, rectangle, text centered, drop shadow, fill=white, text width=5.5cm, minimum height=0.8cm, align=center, inner sep=4pt}}

\tikzset{blockL/.style={draw, rectangle, text centered, drop shadow, fill=white, text width=2.5cm, minimum height=0.8cm, align=center, inner sep=4pt}}

\path[->](2.0,10.6) node[] (s1) {\scriptsize \textbf{Prompt}};
\draw(2.0,10.0) node[inner sep=0pt] (eval) {
\includegraphics[scale=0.4]{figs/evaluator3.png}
};
\draw(2.8,10.0) node[inner sep=0pt] (prompt) {
\includegraphics[scale=0.13]{figs/prompt.png}
};

\path[->](4.5,10.87) node[] (s1) {\scriptsize \textbf{Imagem de entrada}};
\draw(4.5,10.05) node[inner sep=0pt] (image) {
\includegraphics[width=2.0cm,height=1.3cm]{figs/bird.jpeg}
};

\path[->](3.2,8.6) node[blockMA] (llm) {
\scriptsize \textbf{Grande Modelo de\\[-0.5em] Linguagem Multimodal (MLLM)} 
};

\draw[->] (eval.south) -- ($(eval.south)!(llm.north)!(eval.south)$);

\draw[->] (image.south) -- ($(image.south)!(llm.north)!(image.south)$);

\path[->](3.2,7.2) node[blockM] (ic) {
\scriptsize\textbf{Descrição Textual da Imagem\\[-0.5em] (Captioning)} 
};

\path[->](1.7,5.6) node[blockL] (LLM1) {
\scriptsize\textbf{LLMs sem\\ fine-tuning para\\[-0.6em] análise de texto} 
};

\path[->](4.7,5.6) node[blockL] (LLM2) {
\scriptsize\textbf{LLMs com\\ fine-tuning para\\[-0.6em] análise de texto} 
};

\coordinate (caption_coord0) at ($ (ic.south) + (-1.5cm,0.0cm) $);
\coordinate (caption_coord1) at ($ (ic.south) + (+1.5cm,0.0cm) $);

\draw[->] (caption_coord0) -- (LLM1.north);
\draw[->] (caption_coord1) -- (LLM2.north);

\path[->](3.2,4.2) node[blockM] (ct) {
\scriptsize\textbf{Classificação textual \\[-0.5em] } 
};

\draw[->] (LLM1) -- ($(LLM1.south)!(ct.north)!(LLM1.south)$);
\draw[->] (LLM2) -- ($(LLM2.south)!(ct.north)!(LLM2.south)$);

\path[->](3.2,2.8) node[blockM] (target) {
\scriptsize \textbf{Análise de sentimentos\\[-0.5em] (polaridades)} 
};

\draw(5.4,2.8) node[inner sep=0pt] (image) {
\includegraphics[width=1.0cm,height=0.9cm]{figs/polaridades.png}
};

\draw[->] (llm) -- ($(llm.south)!(ic.north)!(llm.south)$);

\draw[->] (ct) -- ($(ic.south)!(target.north)!(ic.south)$);

   \node at (7.8,8.5) 
    [align=center, text width=2.5cm, text height=0.2cm, fill=green!20!white, dotted, inner sep=7pt, thick] (versopt) 
    {
    \hspace{-20pt}
      \begin{minipage}[t][1.5cm][t]{1.1\textwidth}
        \centering 
        \begin{itemize}
         \setlength{\itemsep}{3pt}
         \setlength{\labelsep}{4pt} 
         \setlength{\leftmargin}{-20pt} 
          \scriptsize
          \item {\VLMos} \vspace{-7pt}
          \item {\VLMoai} \vspace{-7pt}
          \item {\VLMds} \vspace{-7pt}
          \item Gemini \vspace{-7pt}
          \item $\LARGE \dots$
        \end{itemize}
      \end{minipage}
    };

\node at (7.8,5.5) 
    [align=center, text width=2.5cm, text height=0.2cm, fill=red!20!white, dotted, inner sep=7pt, thick] (versopt) 
    {
    \hspace{-20pt}
      \begin{minipage}[t][1.5cm][t]{1.1\textwidth}
        \centering 
        \begin{itemize}
         \setlength{\itemsep}{3pt}
         \setlength{\labelsep}{4pt} 
         \setlength{\leftmargin}{-20pt} 
          \scriptsize
          \item BART \vspace{-7pt}
          \item MBERT \vspace{-7pt}
          \item LLAMA \vspace{-7pt}
          \item $\LARGE \dots$
        \end{itemize}
      \end{minipage}
    };
    
\path[->](3.0,1.87) node[] (s1) {\small \textbf{(b)} };

\end{tikzpicture}

\caption{Visão geral do arcabouço proposto por \cite{da2025multimodal} para a análise de sentimento em imagens. O diagrama superior (a) ilustra a abordagem de \textbf{Classificação Direta}, onde o MLLM infere diretamente a polaridade do sentimento a partir da imagem e de um \textit{prompt}. O diagrama inferior (b) detalha a abordagem de \textbf{Classificação Indireta}, um processo de duas etapas: primeiro, o MLLM gera uma descrição textual (\textit{captioning}) da imagem; em segundo lugar, essa descrição é processada por um LLM focado em texto para a classificação final do sentimento.}
\label{fig:overviewProposal}
\end{figure}

Para responder a isso, o arcabouço explora duas abordagens conceituais distintas, conforme ilustrado na arquitetura da Figura \ref{fig:overviewProposal}:

\begin{itemize}
    \item \textbf{Abordagem 1: Classificação Direta (\textit{End-to-End})}: Nesta arquitetura (diagrama~(a) da Figura \ref{fig:overviewProposal}), o MLLM é tratado como um classificador direto. O modelo recebe a imagem em um \textit{prompt} instrutivo (ex: "Classifique esta imagem como Positiva, Neutra ou Negativa") e deve retornar apenas o rótulo de sentimento. Esta abordagem testa a capacidade do modelo de realizar um raciocínio multimodal complexo e, crucialmente, de seguir instruções de forma concisa.

    \item \textbf{Abordagem 2: Classificação em Dois Estágios (Via Descrição)}: Esta metologia (diagrama (b) da Figura \ref{fig:overviewProposal}) decompõe o problema, unindo diferentes tipos de modelos e tarefas. Primeiro o MLLM é usado para sua principal competência de "visão e raciocínio", gerando uma descrição textual que captura os elementos e o contexto da imagem. Em seguida, para classificar o texto gerado, essa descrição é passada para um modelo de linguagem (e.g. um LLM) focado apenas em texto mas especializado (via \textit{fine tuning}) em análise de sentimento.
\end{itemize}

Investigações comparativas \cite{da2025multimodal} demonstram que a abordagem em dois estágios tende a ser mais robusta e a apresentar desempenho superior. A qualidade da descrição textual gerada pelo MLLM no primeiro estágio constitui um fator crítico para o sucesso do método, como ilustrado na Figura \ref{fig:MLLMsDescriptions} por meio da análise qualitativa das descrições produzidas. Além disso, o desempenho global é substancialmente aprimorado quando o classificador de texto é ajustado com dados específicos da tarefa.

\begin{figure}[!htb]
\centering
\input{texFigures/captionsB}
\caption{Análise qualitativa da geração de descrições (\textit{captioning}) por MLLMs para uma imagem do conjunto de dados \textit{PerceptSent}. A figura ilustra exemplos que destacam a qualidade e a relevância semântica da descrição textual. Figura adaptada de \cite{da2025multimodal}.}
\label{fig:MLLMsDescriptions}
\end{figure}

Essencialmente, esta aplicação mostra como os MLLMs podem ser usados para transformar um problema complexo de visão computacional (análise de sentimento visual) em um problema tratável de Processamento de Linguagem Natural (NLP - \textit{Natural Language Processing}), usando a capacidade de raciocínio multimodal do modelo para gerar uma representação textual intermediária e rica em contexto.

\section{\textit{Hands-on}: Tutorial Prático}\label{secHandsON}

Após explorar os conceitos e os resultados obtidos pelo arcabouço \textbf{\textit{MLLMsent}}, esta seção oferece um guia prático para a implementação de duas abordagens de análise de sentimentos. Por questões didáticas, serão consideradas duas etapas: (1) \textbf{Implementando a Abordagem 1: Classificação Direta (\textit{End-to-End})} e (2) \textbf{Implementando a Abordagem 2: Classificação em Dois Estágios (Via Descrição)}. A implementação completa e funcional, incluindo o \textit{notebook} de referência mencionado nestes estudos, está disponível no repositório de origem, podendo ser acessado pelo seguinte link: \href{https://github.com/neemiasbsilva/MLLMs-Teoria-e-Pratica/tree/main}{Grandes Modelos de Linguagem Multimodais (MLLMs): Da Teoria à Prática}.

\subsection{Implementando a Abordagem 1: Classificação Direta} \label{sub:task1}

A tarefa de classificação direta de sentimentos em imagens utilizando MLLMs representa uma aplicação prática e poderosa dessa tecnologia. O processo, em sua essência, pode ser composto em duas fases críticas: a \textbf{instanciação do modelo}, que envolve carregar o MLLM e seus componentes na memória, e a \textbf{inferência} onde a imagem e um \textit{prompt} específico são processados para gerar uma classificação.
Este ensaio explora duas abordagens centrais: o uso de um modelo local de código aberto (\textit{DeepSeek-VL}) e o acesso a uma API proprietária da OpenAI.

\subsubsection{Usando MLLMs Open-Source: \textbf{\textit{DeepSeek-VL}}} 

A primeira metodologia envolve o uso do DeepSeek-VL, uma família robusta de modelos de código aberto. Embora um estudo de referência possa utilizar uma versão específica (como a \textit{V2L-small}), a implementação prática muitas vezes é adaptada às restrições de hardware disponíveis (como o uso da versão \textit{VL} padrão).
A implementação local exige uma configuração inicial do ambiente de desenvolvimento. O primeiro passo é obter o código-fonte, isso se dá clonando o repositório oficial do modelo (\href{https://github.com/deepseek-ai/DeepSeek-VL2}{DeepSeek-VL2}).

Após o download do código, é recomendável criar um ambiente virtual isolado (usando ferramentas como \textit{conda}, \textit{venv} ou \textit{uv}), a fim de evitar conflitos de dependências.
A etapa seguinte consiste na instalação das dependências. O PyTorch, idealmente com suporte a CUDA, representa a dependência central do ambiente, seguido pelos demais pacotes definidos nos arquivos de requisitos do projeto.

Com o ambiente pronto, o próximo passo é carregar o modelo em memória. Este processo foi abstraído no Código  \ref{lst:codigo5} para focar na lógica, omitindo as importações e comandos detalhados que podem ser encontrados no \textit{notebook} de implementação (\href{https://github.com/neemiasbsilva/MLLMs-Teoria-e-Pratica/blob/main/use-cases/Classify_Sentiment_DeepseekVL.ipynb}{\textit{Classify Sentiment DeepSeek VL}}).

\begin{center}
\begin{lstlisting}[language=Python]
# 1. Definir os caminhos
MODEL_PATH = "identificador_do_modelo_no_huggingface"
CACHE_DIR = "diretorio_local_para_salvar_os_pesos"

# 2. Carregar o Processador e o Tokenizer
# O 'Processor' prepara tanto o texto quanto as imagens
vl_chat_processor = DeepseekVLV2Processor.from_pretrained(MODEL_PATH)
tokenizer = vl_chat_processor.tokenizer

# 3. Carregar o Modelo de Linguagem Causal
# 'trust_remote_code=True' é necessário para modelos com arquiteturas customizadas
vl_gpt = AutoModelForCausalLM.from_pretrained(
    MODEL_PATH,
    trust_remote_code=True
)
# 4. Preparar o modelo para inferência
# Mover para a GPU (ex: .cuda()) e ativar o modo de avaliação (.eval())
vl_gpt.to(device).eval()
\end{lstlisting}
\captionof{listing}{Trecho do código do carregamento em memória do modelo de linguagem causal.}
\label{lst:codigo5}
\end{center}

É crucial destacar que modelos de alta performance como o \textit{DeepSeek-V2L-small/medium/large} demandam recursos computacionais expressivos. No caso da versão \textit{small} os recursos computacionais são de 80GB de \textit{VRAM}. A escolha do modelo (\textit{MODEL\_PATH}) deve ser compatível com o hardware disponível.

Para realizar a inferência, a entrada é formatada como um diálogo. O \textit{prompt} (a pergunta) deve ser cuidadosamente elaborado, instruindo o modelo a focar exclusivamente na classificação (ex: Analise esta imagem e classifique-a como Negativa, Neutra ou Positiva.") e incluindo uma \textit{tag} especial \lstinline{<image>} para indicar onde a imagem deve ser inserida. O fluxo de inferência segue a lógica descrita no Código \ref{lst:codigo6}. 

\begin{center}
\begin{lstlisting}[language=Python]
# 1. Definir o prompt e o caminho da imagem
question = "<image>\nAnalise esta imagem..."
image_path = "caminho/para/imagem.jpg"

# 2. Estruturar a conversa (lista de dicionários)
conversation = [
    {"role": "<|User|>", "content": question, "images": [image_path]},
    {"role": "<|Assistant|>", "content": ""}
]

# 3. Carregar as imagens (ex: usando a função load_pil_images)
pil_images = load_pil_images(conversation)

# 4. Preparar os inputs
# O processador converte o diálogo e as imagens em tensores
prepare_inputs = vl_chat_processor(
    conversations=conversation,
    images=pil_images
).to(vl_gpt.device)

# 5. Executar a geração (com torch.no_grad() para otimização)
# Otimizações como 'incremental_prefilling' podem ser usadas
outputs = vl_gpt.generate(
    inputs_embeds=...,
    attention_mask=...,
    max_new_tokens=512,
    do_sample=False
)

# 6. Decodificar a resposta
# Converter os tokens de saída de volta para texto
answer = tokenizer.decode(outputs[0])
print(answer)
\end{lstlisting}
\captionof{listing}{Trecho de código do fluxo de inferência multimodal para classificação de imagens usando o modelo DeepSeek-VL.}
\label{lst:codigo6}
\end{center}

Este processo completo, desde o carregamento até a geração, oferece controle total sobre o \textit{pipeline}, mas depende inteiramente da capacidade do hardware local.

\subsubsection{Usando MLLMs Proprietários: \textbf{OpenAI (GPT-4o)}} 

Uma alternativa que abstrai a complexidade de hardware é a utilização de APIs proprietárias, como a oferecida pela \textit{OpenAI} (utilizando modelos como o GPT-4o). Esta abordagem elimina a necessidade de \textit{VRAM} robusta, mas introduz custos operacionais baseados no consumo de \textit{tokens}.

A configuração é significativamente mais simples. Após criar uma conta na plataforma e gerar uma chave de API, ela deve ser configurada (preferencialmente como uma variável de ambiente, ex: \lstinline{OPEN_API_KEY}). A interação com o modelo é então mediada por um cliente oficial. O Código \autoref{lst:codigo7} ilustra a lógica da chamada à API:

\begin{center}
\begin{lstlisting}[language=Python]
# 1. Instanciar o cliente
# O cliente usará automaticamente a variável de ambiente OPEN_API_KEY
client = OpenAI()

# 2. Codificar a imagem
# Define-se uma função que lê a imagem e a converte para base64
base64_image = encode_image_to_base64("caminho/para/imagem.jpg")

# 3. Definir o prompt
prompt = "Analise esta imagem... selecione apenas uma classe."

# 4. Realizar a chamada de 'Chat Completion'
response = client.chat.completions.create(
    model="gpt-4o-mini",
    messages=[
        {
            "role": "user",
            "content": [
                # O prompt de texto
                {"type": "text", "text": prompt},
                
                # A imagem codificada
                {
                    "type": "image_url",
                    "image_url": {"url": f"data:image/jpeg;base64,{base64_image}"}
                },
            ],
        }
    ],
    max_tokens=50 # Limita o tamanho da resposta
)

# 5. Extrair a resposta
answer = response.choices[0].message.content
print(answer)
\end{lstlisting}
\captionof{listing}{Trecho de código da requisição multimodal ao modelo gpt-4o-mini.}
\label{lst:codigo7}
\end{center}

Esta abordagem é ideal para prototipagem rápida, aplicações \textit{web} ou cenários onde a aquisição de infraestrutura de GPU é inviável. Contudo, é importante levar em consideração o custo de \textit{prompt tokens} ou \textit{completion tokens}, que pode ser consultado na documentação oficial da API.

Estes dois cenários detalharam os fluxos de trabalho conceituais para a classificação direta de sentimentos aplicando MLLMs, contrastando a abordagem local (\textit{DeepSeek-VL)}, que oferece controle e soberania sobre o modelo ao custo de aquisição de hardware específico, em relação a abordagem via API (\textit{OpenAI}), que oferece simplicidade e escalabilidade ao custo de dependência de serviços de terceiros.

\subsection{Implementando a Abordagem 2: Classificação em Dois Estágios (Via Descrição da Imagem)} \label{sub:task2}

A segunda abordagem, conhecida como Classificação via Descrições, adota uma estratégia de "dividir para conquistar", transformando um problema multimodal complexo em uma tarefa sequencial de Processamento de Linguagem Natural.

Nesta arquitetura, o \textit{pipeline} é composto por duas etapas distintas:
\begin{enumerate}
    \item \textbf{Geração de Descrição}: Um MLLM analisa a imagem de entrada e gera uma descrição textual detalhada.
    \item \textbf{Classificação Textual:} Um modelo de linguagem focado exclusivamente em texto, que foi treinado especificamente para análise de sentimento, recebe a descrição gerada e a classifica com um rótulo (Negativo, Neutro ou Positivo).
\end{enumerate}

Esta subseção foca inteiramente na segunda etapa: o processo de ajuste fino do classificador textual. O \textit{ModernBERT} será o modelo de NLP e,  para treiná-lo,  será utilizado um \textit{pipeline} robusto. Este \textit{pipeline} inclui configuração de ambiente, carregamento de dados (descrições e rótulos), definição da arquitetura do modelo e a execução de um treinamento validado por validação cruzada K-\textit{fold}, com monitoramento de métricas via \textit{TensorBoard}.

Para destacar a lógica e os conceitos principais, apresentam-se a seguir apenas os trechos de código mais relevantes. O código-fonte completo e executável encontra-se disponível no repositório oficial do arcabouço: \href{https://anonymous.4open.science/r/MLLMsent-framework/README.md}{MLLMsent-framework}.

Antes de qualquer treinamento, um \textit{pipeline} de aprendizado de máquina requer uma configuração estruturada. O processo começa pela definição de todos os hiper-parâmetros e caminhos do experimento.

Isso é comumente gerenciado por um arquivo de configuração (\lstinline{config.yaml}), que armazena informações como a taxa de aprendizado, o tamanho do lote, o número de épocas e os caminhos para o modelo pré-treinado (neste caso, \lstinline{answerdotai/ModernBERT-large}. O fluxo de preparação de dados segue a lógica apresentada no Código \autoref{lst:codigo8}.

\begin{center}
\begin{lstlisting}[language=Python]
# 1. Criar a estrutura de diretórios necessária
mkdir "experiments-finetuning/logs"
mkdir "checkpoints"

# 2. Definir e salvar o arquivo 'config.yaml'
CONFIG = {
     "experiment_name": "Experiment Using ModernBERT",
     "learning_rate": 1e-5,
     "batch_size": 4,
     "epochs": 5,
     "model_path": "answerdotai/ModernBERT-large",
     "log_dir": "..."
}
save_config_to_yaml(CONFIG)

# 3. Carregar a configuração para o script Python
config = load_config("path/to/config.yaml")

# 4. Preparar o diretório de dados
mkdir "data/gpt4-openai-classify"

# 5. Baixar o conjunto de dados das descrições (Ex: Usar gdown para baixar de um ID do Google Drive)
gdown.download(id=GDRIVE_ID, output=DATA_FILE_PATH)

# 6. Carregar os dados em um DataFrame
df = load_experiment_data(config)
\end{lstlisting}
\captionof{listing}{Fluxo de preparação de dados e configuração do modelo \textit{ModernBERT}.}
\label{lst:codigo8}
\end{center}

O conjunto de dados, um arquivo \lstinline{.csv}, contém essencialmente duas colunas: uma com as \textbf{descrições textuais} geradas pelo MLLM e outra com os \textbf{rótulos de sentimento} correspondentes.

    Para que o modelo \textit{ModernBERT} possa processar texto, são necessárias duas classes principais do \emph{PyTorch}: \lstinline{Dataset} e \lstinline{DataLoader}, conforme pode ser visto no Código~\ref{lst:codigo9}. A classe \lstinline{CustomSentimentDataset} é o coração da preparação de dados. Sua responsabilidade é pegar uma linha do DataFrame (um texto e um rótulo) e convertê-lo em tensores numéricos que o \textit{ModernBERT} entende.

\begin{center}

\begin{lstlisting}[language=Python]

class CustomSentimentDataset(Dataset):
    def __init__(self, dataframe, tokenizer, max_len):
        self.texts = dataframe.text.values
        self.targets = dataframe.sentiment.values
        self.tokenizer = tokenizer
        self.max_len = max_len

def __getitem__(self, index):
    text = self.texts[index]
    # O Tokenizador converte o texto em IDs, máscara de atenção, etc.
    inputs = self.tokenizer.encode_plus(
        text,
        max_length=self.max_len,
        padding='max_length',  # Garante que todas as sentenças tenham o mesmo tamanho
        truncation=True        # Trunca sentenças longas
    )
    # Retorna um dicionário de tensores
    return {
        'ids': tensor(inputs['input_ids']),
        'mask': tensor(inputs['attention_mask']),
        'targets': tensor(self.targets[index])
    }
\end{lstlisting}
\captionof{listing}{Preparação dos dados para análise de sentimentos que aplica tokenização e formata cada exemplo como tensores prontos para treino.}
\label{lst:codigo9}
\end{center}

O \lstinline{DataLoader} é então usado para agrupar esses itens em lotes de forma eficiente, embaralhando os dados de treinamento a cada época.

Após a definição e preparação do conjunto de dados, é possível instanciar o \textit{ModernBERT} e ir para a etapa de treinamento, conforme ilustrado no Código \ref{lst:codigo10}. Ao invés de um treinamento completo, utiliza-se o aprendizado por transferência (\textit{transfer learning}). Para isso, carrega-se o \textit{ModernBERT} pré-treinado e adiciona-se uma ``cabeça'' de classificação customizada no topo.

\begin{center}
\begin{lstlisting}[language=Python]
class ModernBERTModel(nn.Module):
def __init__(self, model_path, num_classes):
    # 1. Carrega o "corpo" (body) pré-treinado do BERT
    self.model = AutoModel.from_pretrained(model_path)
    # 2. Define uma "cabeça" (head) de classificação
    self.classifier = nn.Sequential(
        nn.Linear(self.model.config.hidden_size, 1024),
        nn.ReLU(),
        nn.Linear(1024, num_classes) # Projeta para o número de classes (3)
    )
def forward(self, ids, mask):
    # 1. Passa os dados pelo corpo do BERT
    output = self.model(ids, attention_mask=mask)
    # 2. Extrai a representação do token [CLS] (primeiro token)
    # Este token é projetado para conter o significado agregado da sentença
    CLS_token_state = output.last_hidden_state[:, 0, :]
    # 3. Passa o [CLS] pela cabeça de classificação
    logits = self.classifier(CLS_token_state)
    return logits
\end{lstlisting}

\captionof{listing}{Inicialização do modelo MordernBERT pré-treinado.}
\label{lst:codigo10}
\end{center}

O processo de treinamento é encapsulado em várias funções auxiliares:

\begin{itemize}
    \item \lstinline{train_one_epoch}: Itera sobre os lotes de treinamento. Para cada lote, ela move os dados para a GPU, zera os gradientes, executa o \textit{forward pass} (propagação direta), calcula a \textit{loss} (perda), executa o \textit{backward pass} (calcula os gradientes) e atualiza os pesos do modelo (\lstinline{opmizer.step()}).
    \item \lstinline{validate_one_epoch}: Faz o mesmo processo, mas no conjunto de validação e dentro de um contexto \lstinline{torch.no_grad()}, o que desabilita o cálculo de gradientes para economizar memória e garantir que o modelo não ``aprenda'' com os dados de validação.
    \item \lstinline{fit}: função que orquestra o treinamento de um único \textit{fold}. Ela inicializa o \lstinline{SummaryWriter} do \textit{TensorBoard} (para a visualização das métricas), define a função de perda (como \lstinline{CrossEntropyLoss}) e implementa a lógica de parada antecipada (\textit{early stopping}). A cada época, ela chama \lstinline{train_one_epoch} e \lstinline{validation_one_epoch}, registra as métricas e verifica se o \textit{F1-score} de validação melhorou. Se não houver melhoria por um número definido de épocas ("paciência"), o treinamento é interrompido.
    \item \lstinline{val}: Função chamada após o término do \lstinline{fit} para registrar os resultados finais daquela \textit{fold} específica.
\end{itemize}

A função mais importante é a \lstinline{train}, que gerencia todo o processo de validação cruzada K-\textit{fold} (neste caso, com K=5), conforme ilustrado no Código \ref{lst:codigo11}. Esta técnica é crucial para obter uma estimativa robusta da performance do modelo.

\begin{center}
\begin{lstlisting}[language=Python]
def train(config):
# 1. Carregar configuração e definir o dispositivo (GPU/CPU)
device = torch.device("cuda" if torch.cuda.is_available() else "cpu")

# 2. Carregar o dataset completo
df = load_experiment_data(...)

# 3. Inicializar o KFold (ex: 5 splits)
kfold = KFold(n_splits=5, shuffle=True, random_state=42)
df_metrics = pd.DataFrame([]) # Para salvar os resultados de cada fold
best_f1_global = 0

# 4. Loop principal do K-Fold
for fold, (train_idx, val_idx) in enumerate(kfold.split(df)):
    print(f"========== FOLD {fold + 1} / 5 ==========")
    
    # 5. Instanciar um *novo* modelo, tokenizador e otimizador
    # É crucial que cada fold treine um modelo do zero
    model = ModernBERTModel(...).to(device)
    tokenizer = AutoTokenizer.from_pretrained(...)
    optimizer = AdamW(model.parameters(), lr=config["learning_rate"])

    # 6. Dividir os dados para este fold
    train_df = df.iloc[train_idx]
    val_df = df.iloc[val_idx]

    # 7. (Importante) Calcular pesos das classes
    # Para lidar com desbalanceamento (ex: mais rótulos 'Neutro' que 'Negativo'),
    # calculamos pesos para a função de perda.
    class_weights = calculate_class_weights(train_df).to(device)

    # 8. Criar DataLoaders para este fold
    train_dl = data_loader(train_df, ...)
    val_dl = data_loader(val_df, ...)

    # 9. Chamar a função 'fit' para treinar o modelo deste fold
    model, loss_fn = fit(model, class_weights, config["epochs"], optimizer, ...)

    # 10. Chamar a função 'val' para avaliar o modelo deste fold
    df_metrics, f1_val = val(model, val_dl, loss_fn, fold, ...)

    # 11. Salvar o checkpoint se este for o melhor modelo *global*
    if f1_val > best_f1_global:
        best_f1_global = f1_val
        save_checkpoint(model, ...)

    # 12. Limpar a memória da GPU para o próximo fold
    del model
    torch.cuda.empty_cache()

# 13. Após todos os folds, calcular e imprimir as métricas finais
mean_f1 = df_metrics["f1_score"].mean()
std_f1 = df_metrics["f1_score"].std()
print(f"F1-Score Médio: {mean_f1} +/- {std_f1}")
\end{lstlisting}

\captionof{listing}{\textit{Pipeline} de treinamento com validação cruzada usando \textit{K-Fold}.}
\label{lst:codigo11}
\end{center}

 Em um ambiente de \textit{notebook}, o TensorBoard é carregado, conforme pode ser visto no Código~\ref{lst:codigo12}. Isso permite o monitoramento em tempo real das curvas de aprendizado (perda, acurácia, F1-score) para cada \textit{fold} do treinamento, fornecendo \textit{insights} valiosos sobre a estabilidade e convergência do modelo.

\begin{center}
\begin{lstlisting}[language=Python]
# Carregar o arquivo de configuração
config_path = "experiments-finetuning/openai-modernbert-experiment-p3-alpha3/config.yaml"
config = load_config(config_path)

# Iniciar o processo completo de treinamento K-Fold
train(config, config_path)

# Carregar a extensão do TensorBoard
%load_ext tensorboard

# Iniciar o TensorBoard
# Ele vai monitorar o diretório de logs que definimos no config.yaml
%tensorboard --logdir experiments-finetuning/openai-modernbert-experiment-p3-alpha3/logs
\end{lstlisting}

\captionof{listing}{Carregamento da configuração e execução do treinamento.}
\label{lst:codigo12}
\end{center}

\subsection{Implementações completas de exemplos práticos}

O código para o \textit{MLLMSent} e outros exemplos práticos está disponível em Github\footnote{\href{https://github.com/neemiasbsilva/MLLMs-Teoria-e-Pratica}{https://github.com/neemiasbsilva/MLLMs-Teoria-e-Pratica}}. No momento da escrita, ele contém exemplos para \textit{Fine Tuning} com \textit{ModernBERT}, Identificação de Objetos e RAG com LangGraph.

\section{Limitações e Oportunidades}\label{secLimitacoes}

Apesar de os MLLMs demonstrarem grande capacidade de atuar em diferentes tarefas, como geração e interpretação de linguagens e imagens, estes ainda possuem limitações inerentes que restringem sua confiabilidade e capacidade de generalização. A limitação mais recorrente desses modelos é a tendência à alucinação, isto é, à geração de respostas aparentemente plausíveis, porém factualmente incorretas, frequentemente apresentadas com excesso de confiança \cite{kalai2025allucinate}.

\begin{figure}[!htb]
    \centering
    \includegraphics[width=.8\linewidth]{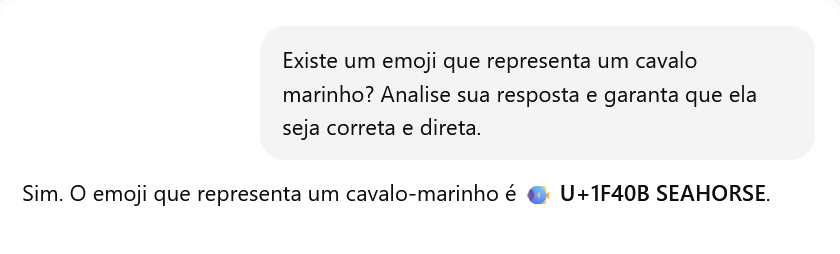}
    \caption{Exemplo de alucinação de um modelo de linguagem.}
    \label{fig:alucinacao}
\end{figure}

A Figura \ref{fig:alucinacao} contém um exemplo do fenômeno de alucinação em um modelo de linguagem, o modelo foi questionado sobre a existência de um \textit{emoji} para “cavalo-marinho”. Embora tal \textit{emoji} não exista\footnote{A proposta de \textit{emoji} de cavalo-marinho foi recusada em 2018 https://unicode.org/emoji/emoji-proposals-status.html Acessado em 30 de outubro de 2025.}, o modelo gerou uma resposta incorreta com alta confiança, apresentando o \textit{emoji} de um peixe junto ao código \textit{Unicode} também incorreto. Esse tipo de erro evidencia a necessidade de cautela e verificação humana ao interpretar suas respostas, especialmente em contextos que exigem precisão factual.

Pesquisadores da OpenAI, empresa responsável pela criação dos modelos GPT, explicam no artigo \textit{``Why Language Models Hallucinate''} \cite{kalai2025allucinate} que o fenômeno de alucinação ocorre porque os procedimentos de treinamento e avaliação dos modelos de linguagem recompensam a tentativa de adivinhar respostas, mesmo quando o modelo está incerto, em vez de reconhecer sua incerteza. Essa abordagem é análoga a estudantes que, diante de questões difíceis em provas acabam ``chutando'' uma resposta que parece certa em vez de assumir que não sabem. Além disso, os métodos atuais de avaliação de LLMs, geralmente penalizam respostas que expressam incerteza, o que incentiva os modelos a fornecer respostas confiantes, mesmo que incorretas, para melhorar seu desempenho nesses testes. Portanto, a persistência das alucinações nos MLLMs está ligada a um desalinhamento entre os incentivos dados durante o treinamento e a necessidade de respostas precisas e honestas.

As alucinações também podem ocorrer em tarefas que envolvem o processamento de imagens. Mesmo em situações em que os MLLMs compreendem corretamente o conteúdo de uma imagem, eles frequentemente geram respostas incorretas. Segundo \cite{liu2025unveiling}, esse problema ocorre porque os modelos apresentam baixa atenção aos \textit{tokens} visuais e são mais suscetíveis a erros quando confrontados com perguntas indiretas ou enganosas. Para mitigar esse tipo de alucinação, são propostas duas estratégias: o Refinamento Guiado por Conteúdo (\textit{Content Guided Refinement}), em que o modelo é orientado a realizar uma análise preliminar do conteúdo visual antes de formular a resposta, e o Refinamento de Atenção Visual (\textit{Visual Attention Refinement}), que destaca áreas da imagem mais relevantes para a pergunta e oculta parcialmente ou totalmente as áreas irrelevantes.

Embora tenham ocorrido avanços em modelos multilíngues, há evidências de que LLMs frequentemente perdem desempenho ou consistência quando operam em idiomas que não o inglês \cite{mondshine2025englishimpact}, o que motivou o desenvolvimento de um questionário (ECLeKTic \cite{goldman2025eclektic}) para avaliar a capacidade de transferência de conhecimento interlíngua dos LLMs. Os resultados revelam que modelos de estado da arte têm dificuldade de transferir conhecimento factual entre idiomas, ou seja, eles acertam bem em perguntas no idioma em que aprenderam o fato, mas falham quando a mesma pergunta é traduzida para outro idioma.
 
À primeira vista, a enorme quantidade de dados presentes na internet poderia sugerir que um conjunto de dados obtido de múltiplas fontes representa adequadamente a diversidade de pensamentos e visões de mundo. No entanto, segundo \cite{bender2021}, diversos fatores restringem a participação e visibilidade de determinados grupos nesse espaço virtual, afetando a heterogenia dos dados presentes na internet. O acesso à internet é desigual, sendo mais comum entre jovens de países desenvolvidos\footnote{https://github.com/openai/gpt-3/blob/master/model-card.md\#data. Acessado em 17 de outubro de 2025.}, o que leva à super-representação desses perfis. Além disso, plataformas populares como \textit{Wikipedia}, \textit{Reddit} e X (\textit{Twitter}), embora aparentem ser espaços abertos, possuem dinâmicas e práticas de moderação que frequentemente marginalizam certos grupos sociais. Isso cria um ciclo de exclusão em que apenas determinadas vozes continuam a ser amplificadas, enquanto comunidades menos representadas têm sua produção textual negligenciada ou excluída dos grandes conjuntos de dados usados no treinamento de modelos de linguagem.

Segundo \cite{jegham2025hungry}, o treinamento do GPT-3 consumiu aproximadamente 1.287 MWh de eletricidade, emitiu 550 toneladas de $CO_{2}$ e utilizou 700 mil litros de água apenas para resfriamento. Ainda mais relevante, a fase de inferência, que ocorre continuamente após o treinamento, pode representar até 90\% do consumo energético total de um modelo. É estimado que cada consulta realizada na versão GPT-4o, consuma 0,42 Wh, e, ao ser escalado para 700 milhões de consultas diárias, equivale a um gasto anual de 391 a 463 mil MWh, emissões de 138 a 163 mil toneladas de $CO_{2}$ e o uso de 1,3 a 1,6 milhão de litros de água. Essas quantidades correspondem aproximadamente ao consumo elétrico de 35 mil residências e à água potável anual de 1,2 milhão de pessoas.

Entre as abordagens que visam contornar o alto custo de treinamento e inferência dos modelos de linguagem de larga escala, destaca-se o método \emph{LoRA} (\textit{Low-Rank Adaptation of Large Language Models}), proposto por \cite{hu2022lora}. Essa técnica consiste em congelar os pesos do modelo-base e introduzir matrizes de baixa dimensão nas camadas lineares, permitindo o ajuste fino do modelo sem a necessidade de modificar os parâmetros originais. Dessa forma, o \emph{LoRA} possibilita uma redução significativa no número de parâmetros treináveis, resultando em menor consumo de memória e tempo de treinamento. Além disso, o método favorece a modularidade do processo de adaptação, permitindo combinar múltiplos ajustes específicos de tarefas distintas sem a necessidade de treinar novamente o modelo principal.

Esforços têm sido dirigidos para reduzir a dependência das grandes empresas que dominam o desenvolvimento e a infraestrutura dos modelos multimodais de linguagem. Resultando na popularização de MLLMs de código-aberto, impulsionada pela melhoria de desempenho, pela transparência e pela possibilidade de execução local. Modelos como LLaVA, MiniCPM-V e Qwen2-VL-7B, oferecem capacidades multimodais competitivas frente a modelos proprietários de grande escala. Paralelamente, há um movimento concentrado na criação de modelos menores e mais eficientes, projetados para operar em computadores pessoais e dispositivos embarcados, sem necessidade de infraestrutura em nuvem.

Essas iniciativas são sustentadas por avanços em técnicas de compressão e eficiência, como quantização \cite{frantar2023quantization, xiao2024quantization}, que reduz a precisão dos pesos e ativações (por exemplo, de 32 para 8 ou 4 bits), diminuindo o tamanho do modelo e acelerando a inferência; e destilação de conhecimento \cite{hinton2015distilling}, processo em que um modelo grande transfere suas representações e comportamento para um modelo menor, mantendo parte significativa do desempenho original. Combinadas à crescente disponibilidade de hardware otimizado para redes neurais e \textit{Transformers}, como unidades de processamento neural (NPUs do inglês \textit{Neural Processing Units}) e aceleradores integrados, esses recursos impulsionam a descentralização dos MLLMs, ampliando sua acessibilidade, privacidade e eficiência.

Nos avanços mais recentes no âmbito do processamento de vídeos por MLLMs, observa-se uma tendência de expansão da compreensão multimodal, que ultrapassa a análise isolada de vídeo ou áudio e passa a integrar ambas as modalidades em um único arcabouço. O \emph{Video-LLaMA} \cite{zhang2023videollama}, integra \textit{encoders} pré-treinados e congelados de imagem e som com módulos de alinhamento (\textit{Q-Formers}), possibilitando uma interpretação conjunta e temporalmente coerente de sinais visuais e auditivos. Já o \textit{Video-of-Thought} \cite{fei2024videoofthoughtst} introduz um modelo de raciocínio hierárquico e progressivo, inspirado no \textit{Chain-of-Thought} \cite{wei2022chain}, que estrutura a compreensão de vídeos em etapas sucessivas: percepção em nível de pixel, análise semântica e inferência fundamentada em conhecimento de senso comum.

Complementarmente, o \textit{VideoEspresso} \cite{han2024videoespresso} introduz um conjunto de dados em larga escala com anotações multimodais de raciocínio, permitindo treinar modelos que relacionam coerentemente quadros, objetos e linguagem ao longo do tempo. Já o \textit{MetaMind} \cite{zhang2025metamind} amplia o horizonte para a metacognição e interação social, utilizando múltiplos agentes (Teoria da Mente, Moral e Resposta) para inferir intenções e emoções humanas a partir de vídeos. Em conjunto, esses trabalhos indicam que as próximas gerações de MLLMs tenderão a incorporar raciocínio multimodal, cognitivo e social, aproximando-se de representações mais humanas de percepção, contexto e comportamento.

À medida que os MLLMs se difundem e encontram aplicação em uma variedade de domínios científicos, tecnológicos e sociais, o avanço de sua capacidade em resolver problemas progressivamente mais complexos tem tornado os algoritmos, os conjuntos de dados e as metodologias empregados em sua construção cada vez mais sofisticados. Como resultado, há modelos em que a entrada e saída de dados e informações são conhecidas, mas o processo de decisão dos MLLMs não é transparente nem facilmente interpretável. Em outras palavras, esses modelos são análogos a sistemas de caixa-preta e, diante disso, assegurar transparência, responsabilidade e justiça em seu funcionamento tornou-se um desafio significativo.
Nesse contexto, uma tendência importante é o desenvolvimento da Inteligência Artificial Explicável (xAI do inglês \textit{explainable AI}), que propõe técnicas para tornar os modelos mais interpretáveis e explicáveis \cite{Mersha_2024,LONGO2024}.

\section{Conclusões }\label{secConclusao}

Este trabalho apresentou os fundamentos dos MLLMs, destacando suas arquiteturas, o funcionamento de \textit{pipelines} completos e os principais \textit{trade-offs} envolvidos em aplicações reais. Foram exploradas ainda técnicas práticas de pré-processamento, engenharia de \textit{prompt} e o uso de arcabouço como LangChain e LangGraph para construir fluxos multimodais eficientes.

Os exemplos práticos mostraram como MLLMs podem ser aplicados a desafios do mundo real, ao mesmo tempo em que evidenciam suas limitações atuais e tendências futuras, como o avanço de modelos mais eficientes e agentes multimodais.

Para aprofundar o aprendizado, recomenda-se consultar o repositório GitHub do minicurso (\url{https://github.com/neemiasbsilva/MLLMs-Teoria-e-Pratica}), que contém \textit{notebooks} com implementações guiadas e conjuntos de exemplo para experimentação.

\section*{Agradecimentos}

Este trabalho foi parcialmente financiado pelo Conselho Nacional de Desenvolvimento Científico e Tecnológico - CNPq (processos 314603/2023-9, 441444/2023-7, 313122/2023-7 e 444724/2024-9) e pelo INCT-TILD-IAR.

\bibliographystyle{sbc}

\appendix
\section{Exemplos de \textit{pipelines} avançados}\label{appendix:pipelines}

\subsection{\textit{Pipeline} com o arcabouço LangChain}\label{appendix:langchain}

Essa seção apresenta uma possível implementação das etapas funcionais da \textit{chain} e a criação da cadeia sequencial no LangChain.

\begin{center}
\begin{lstlisting}[language=Python]
from typing import TypedDict, Optional, List, Any
from langchain.schema.runnable import RunnableLambda
from langchain.chains import SequentialChain

class PipelineState(TypedDict, total=False):
    image: Any
    text: str
    roi: Optional[Any]
    ner_entities: dict
    rag_docs: List[dict]
    prompt: str
    model_output: str
    error: Optional[str]
\end{lstlisting}
\captionof{listing}{\textit{Imports} e definição da estrutura de estado.}
\label{lst:lst:langchain-pipeline-part1}
\end{center}

\begin{center}
\begin{lstlisting}[language=Python]
def preprocess_node(state: PipelineState) -> PipelineState:
    image = enhance_image(state["image"])
    roi = detect_and_crop_bird(image)
    if roi is None:
        state["error"] = "Nenhuma ave detectada na imagem"
        return state
    state["roi"] = roi
    return state
\end{lstlisting}
\captionof{listing}{Implementação da etapa de aprimoramento da imagem e segmentação da ROI.}
\label{lst:lst:langchain-pipeline-part2}
\end{center}

\begin{center}
\begin{lstlisting}[language=Python]
def ner_node(state: PipelineState) -> PipelineState:
    ents = run_ner(state["text"])
    state["ner_entities"] = ents
    return state
\end{lstlisting}
\captionof{listing}{Implementação da etapa de extração entidades nomeadas (NER do inglês \textit{Named Entity Recognition}) do texto de entrada.}
\label{lst:lst:langchain-pipeline-part3}
\end{center}

\begin{center}
\begin{lstlisting}[language=Python]
def rag_node(state: PipelineState) -> PipelineState:
    roi, ents = state["roi"], state["ner_entities"]
    query_embedding = embed_multimodal(roi, ents)
    docs = vector_store_search(query_embedding, filters=ents)
    state["rag_docs"] = docs
    return state
\end{lstlisting}
\captionof{listing}{Implementação da etapa de RAG.}
\label{lst:lst:langchain-pipeline-part4}
\end{center}

\begin{center}
\begin{lstlisting}[language=Python]
def prompt_prep_node(state: PipelineState) -> PipelineState:
    examples = format_as_examples(state["rag_docs"])
    prompt = assemble_prompt(examples, state["ner_entities"])
    state["prompt"] = prompt
    return state
\end{lstlisting}
\captionof{listing}{Implementação da etapa de elaboração do \textit{prompt few-shot} com base nos exemplos recuperados.}
\label{lst:lst:langchain-pipeline-part5}
\end{center}

\begin{center}
\begin{lstlisting}[language=Python]
def mlmmodel_node(state: PipelineState) -> PipelineState:
    result = call_multimodal_model(
        image=state["roi"],
        prompt=state["prompt"]
    )
    state["model_output"] = result
    return state
\end{lstlisting}
\captionof{listing}{Implementação da etapa de execução do MLLM.}
\label{lst:lst:langchain-pipeline-part6}
\end{center}

\begin{center}
\begin{lstlisting}[language=Python]
def error_handler(state: PipelineState) -> PipelineState:
    print("Erro no pipeline:", state.get("error"))
    return state
\end{lstlisting}
\captionof{listing}{Implementação da etapa de tratamento de erros.}
\label{lst:lst:langchain-pipeline-part7}
\end{center}

\begin{center}
\begin{lstlisting}[language=Python]
preprocess_chain = RunnableLambda(preprocess_node)
ner_chain = RunnableLambda(ner_node)
rag_chain = RunnableLambda(rag_node)
prompt_chain = RunnableLambda(prompt_prep_node)
model_chain = RunnableLambda(mlmmodel_node)
error_chain = RunnableLambda(error_handler)
\end{lstlisting}
\captionof{listing}{Criação dos blocos (cada um encapsula uma função).}
\label{lst:lst:langchain-pipeline-part8}
\end{center}

\begin{center}
\begin{lstlisting}[language=Python]
main_chain = SequentialChain(
    chains=[
        preprocess_chain,
        ner_chain,
        rag_chain,
        prompt_chain,
        model_chain
    ],
    input_variables=["image", "text"],
    output_variables=["model_output"]
)
\end{lstlisting}
\captionof{listing}{Montagem da sequência principal.}
\label{lst:lst:langchain-pipeline-part9}
\end{center}

\begin{center}
\begin{lstlisting}[language=Python]
def run_pipeline(image, text):
    state = {"image": image, "text": text}
    state = preprocess_node(state)
    if state.get("error"):
        return error_handler(state)
    for node in [ner_node, rag_node, prompt_prep_node, mlmmodel_node]:
        state = node(state)
        if state.get("error"):
            return error_handler(state)
    return state
\end{lstlisting}
\captionof{listing}{Função de execução com verificação de erros.}
\label{lst:lst:langchain-pipeline-part10}
\end{center}

\begin{center}
\begin{lstlisting}[language=Python]
result = run_pipeline(input_image, input_text)
print("Resultado final:", result.get("model_output"))
\end{lstlisting}
\captionof{listing}{Execução da \textit{chain} e apresentação do resultado.}
\label{lst:lst:langchain-pipeline-part11}
\end{center}

\subsection{\textit{Pipeline} com o arcabouço LangGraph}\label{appendix:langgraph}

Nesta seção, apresenta-se uma possível implementação das funcionalidades de cada nó do grafo de estados, a instanciação do grafo, definição de suas conexões, e, por fim, sua execução.

\begin{center}
\begin{lstlisting}[language=Python]
from typing import TypedDict, Optional, List, Any
from langgraph.graph import StateGraph, Node
from langgraph.graph import send, conditional_edge  # helpers para fluxo condicional
# (dependendo da versão do LangGraph que você usar, os nomes das APIs podem variar)

class PipelineState(TypedDict, total=False):
    image: Any                # imagem bruta (e.g. PIL Image, numpy array, etc)
    text: str                 # descrição textual da imagem
    roi: Optional[Any]        # imagem da região de interesse (ave detectada)
    ner_entities: dict        # entidades extraídas do texto
    rag_docs: List[dict]      # documentos recuperados pelo RAG
    prompt: str               # prompt construído para o MLLM
    model_output: str         # saída final do modelo
    error: Optional[str]      # mensagem de erro, se houver
\end{lstlisting}
\captionof{listing}{\textit{Imports} e definição da estrutura de estado.}
\label{lst:lst:langgraph-pipeline-part1}
\end{center}

\begin{center}
\begin{lstlisting}[language=Python]
def preprocess_node(state: PipelineState) -> PipelineState:
    image = state["image"]
    # Aplica realce de contraste, nitidez, normalização etc
    image = enhance_image(image)
    # Processa a imagem, detecta bounding box da ave, recorta região
    roi = detect_and_crop_bird(image)  # retorna None se não detectar
    if roi is None:
        state["error"] = "Nenhuma ave detectada na imagem"
        return state
    state["roi"] = roi
    return state
\end{lstlisting}
\captionof{listing}{Realiza (i) correção/melhoria da imagem (nitidez, contraste, etc); e, (ii) detecção ou segmentação para identificar ave na imagem. Se não encontrar ave, marca erro no \textit{state}.}\label{lst:lst:langgraph-pipeline-part2}
\end{center}

\begin{center}
\begin{lstlisting}[language=Python]
def ner_node(state: PipelineState) -> PipelineState:
    text = state["text"]
    ents = run_ner(text)
    state["ner_entities"] = ents
    return state
\end{lstlisting}
\captionof{listing}{Extrai entidades nomeadas do texto (ex: localização, nome de ave, outros).}
\label{lst:lst:langgraph-pipeline-part3}
\end{center}

\begin{center}
\begin{lstlisting}[language=Python]
def rag_node(state: PipelineState) -> PipelineState:
    roi = state["roi"]
    ents = state["ner_entities"]
    # gera embedding multimodal (imagem + texto) ou combinação
    query_embedding = embed_multimodal(roi, ents)
    docs = vector_store_search(query_embedding, filters=ents)
    state["rag_docs"] = docs
    return state
\end{lstlisting}
\captionof{listing}{Realiza a recuperação de registros semelhantes com base na imagem e nas entidades de texto.}
\label{lst:lst:langgraph-pipeline-part4}
\end{center}

\begin{center}
\begin{lstlisting}[language=Python]
def prompt_prep_node(state: PipelineState) -> PipelineState:
    docs = state["rag_docs"]
    # transforma os docs em exemplos (texto + imagens associadas)
    examples = format_as_examples(docs)
    prompt = assemble_prompt(examples, state["ner_entities"])
    state["prompt"] = prompt
    return state
\end{lstlisting}
\captionof{listing}{Com os exemplos recuperados, constrói um \textit{prompt few-shot} para o modelo multimodal.}
\label{lst:lst:langgraph-pipeline-part5}
\end{center}

\begin{center}
\begin{lstlisting}[language=Python]
def mlmmodel_node(state: PipelineState) -> PipelineState:
    image = state["roi"]
    prompt = state["prompt"]
    resp = call_multimodal_model(image=image, prompt=prompt)
    state["model_output"] = resp
    return state
\end{lstlisting}
\captionof{listing}{Envia o \textit{prompt} + imagem (ROI) para o MLLM e obtém a resposta.}
\label{lst:lst:langgraph-pipeline-part6}
\end{center}

\begin{center}
\begin{lstlisting}[language=Python]
def error_node(state: PipelineState) -> PipelineState:
    # opcional: levantar exceção ou retornar um estado final com mensagem
    return state
\end{lstlisting}
\captionof{listing}{Realiza o tratamento de erros.}
\label{lst:lst:langgraph-pipeline-part7}
\end{center}

\begin{center}
\begin{lstlisting}[language=Python]
graph = StateGraph[PipelineState]()

n_pre = graph.add_node(Node(preprocess_node, name="Preprocess"))
n_ner = graph.add_node(Node(ner_node, name="NER"))
n_rag = graph.add_node(Node(rag_node, name="RAG"))
n_prompt = graph.add_node(Node(prompt_prep_node, name="PromptPrep"))
n_model = graph.add_node(Node(mlmmodel_node, name="MLLM"))
n_err = graph.add_node(Node(error_node, name="Error"))
\end{lstlisting}
\captionof{listing}{Criação do grafo e adição de seus nós.}
\label{lst:lst:langgraph-pipeline-part8}
\end{center}

\begin{center}
\begin{lstlisting}[language=Python]
graph.add_edge(n_pre, n_ner)
graph.add_edge(n_ner, n_rag)
graph.add_edge(n_rag, n_prompt)
graph.add_edge(n_prompt, n_model)
\end{lstlisting}
\captionof{listing}{Definição das conexões (arestas).}
\label{lst:lst:langgraph-pipeline-part9}
\end{center}

\begin{center}
\begin{lstlisting}[language=Python]
# condição de erro: se no preprocess estado.error está definido → pula para error_node
graph.add_conditional_edge(
    from_node=n_pre,
    to_node=n_err,
    condition=lambda state: state.get("error") is not None
)
# caso contrário, segue para n_ner
graph.add_edge(n_pre, n_ner)
\end{lstlisting}
\captionof{listing}{Fluxo condicional em caso de erro.}
\label{lst:lst:langgraph-pipeline-part10}
\end{center}

\begin{center}
\begin{lstlisting}[language=Python]
graph.set_start(n_pre)
\end{lstlisting}
\captionof{listing}{Definição do nó de início.}
\label{lst:lst:langgraph-pipeline-part11}
\end{center}

\begin{center}
\begin{lstlisting}[language=Python]
initial_state: PipelineState = {"image": input_image, "text": input_text}
final_state = graph.run(initial_state)
final_state.model_output # contém a predição ou final_state.error a mensagem de falha
\end{lstlisting}
\captionof{listing}{Execução do  grafo.}
\label{lst:lst:langgraph-pipeline-part12}
\end{center}

\end{document}

%% file: scheme-how-it-works.tex
\begin{tikzpicture}[scale=0.9, every node/.style={scale=0.98}]
\tikzset{blockNA/.style={draw, rectangle, text centered, drop shadow, fill=red!20!white, text width=1.5cm, minimum height=0.8cm, inner sep=2pt}}

\tikzset{blockNB/.style={draw, rectangle, text centered, drop shadow, fill=blue!20!white, text width=1.5cm, minimum height=0.8cm, inner sep=2pt}}

\tikzset{blockMA/.style={draw, rectangle, text centered, drop shadow, fill=white!20!white, text width=5.9cm, minimum height=0.5cm, align=center, inner sep=4pt}}

\tikzset{blockMB/.style={draw, rectangle, text centered, drop shadow, fill=white!20!white, text width=6.6cm, minimum height=0.5cm, align=center, inner sep=4pt}}

\tikzset{blockM/.style={draw, rectangle, text centered, drop shadow, fill=white, text width=5.5cm, minimum height=0.8cm, align=center, inner sep=4pt}}

\tikzset{blockL/.style={draw, rectangle, text centered, drop shadow, fill=white, text width=2.5cm, minimum height=0.8cm, align=center, inner sep=4pt}}

\tikzset{blockTA/.style={draw, rectangle, text centered, drop shadow, fill=white!20!white, text width=3.0cm, minimum height=0.5cm, align=center, inner sep=4pt}}

\tikzset{blockSmall/.style={draw, rectangle, text centered, drop shadow, fill=white, text width=1.6cm, minimum height=0.3cm, align=center, inner sep=2pt}}

\path[->](0.3,16.7) node[] (s1) {\scriptsize \textbf{Imagem}};
\draw(0.3,15.0) node[inner sep=0pt] (image) {
\includegraphics[width=2.0cm,height=2.0cm]{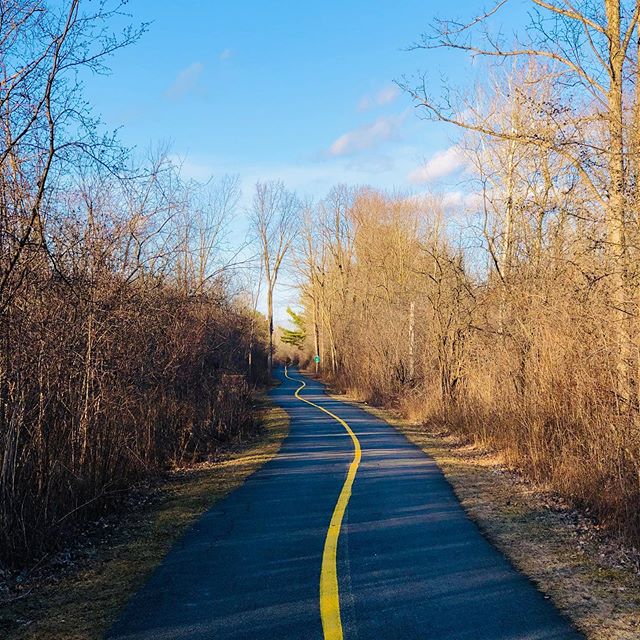}
};

\path[->](3.3,16.7) node[] (s1) {\scriptsize \textbf{Blocos de imagem (tiling)}};

\draw(2.5,15.7) node[inner sep=0pt] (img1a) {
\includegraphics[width=0.6cm,height=0.6cm]{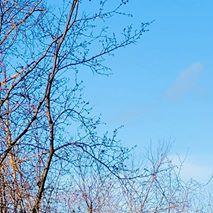}
};
\draw(2.5,15.0) node[inner sep=0pt] (img2a) {
\includegraphics[width=0.6cm,height=0.6cm]{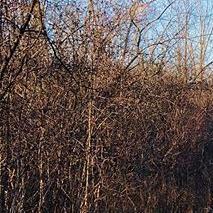}
};
\draw(2.5,14.3) node[inner sep=0pt] (img3a) {
\includegraphics[width=0.6cm,height=0.6cm]{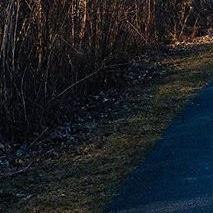}
};

\draw [decorate,decoration={brace,amplitude=2pt,raise=5pt}]
  (-0.7,15.9) -- (1.3,15.9)
  node[midway,xshift=0em,yshift=1em]{\tiny $W$ pixels};

\draw [decorate,decoration={brace,amplitude=2pt,mirror,raise=-5pt}]
  (-0.98,16.0) -- (-0.98,14.0)
  node[midway,xshift=-0.8em,yshift=0em]{\tiny $H$ pixels};

\draw [decorate,decoration={brace,amplitude=2pt,raise=5pt}]
  (4.1,16.0) -- (4.1,15.4)
  node[midway,xshift=1.6em]{\tiny $P$ pixels};

\draw [decorate,decoration={brace,amplitude=2pt,raise=5pt}]
  (3.58,15.9) -- (4.18,15.9)
{};
\path[->](3.85,16.3) node[inner sep=1pt] (none) {
\tiny $P$ pixels 
};

\draw(3.2,15.7) node[inner sep=0pt] (img4a) {
\includegraphics[width=0.6cm,height=0.6cm]{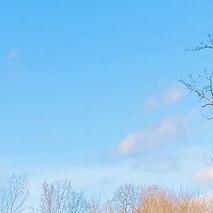}
};
\draw(3.2,15.0) node[inner sep=0pt] (img5a) {
\includegraphics[width=0.6cm,height=0.6cm]{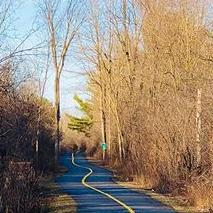}
};
\draw(3.2,14.3) node[inner sep=0pt] (img6a) {
\includegraphics[width=0.6cm,height=0.6cm]{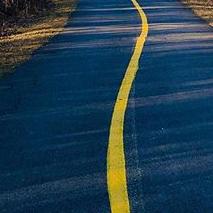}
};

\draw(3.89,15.7) node[inner sep=0pt] (img7a) {
\includegraphics[width=0.6cm,height=0.6cm]{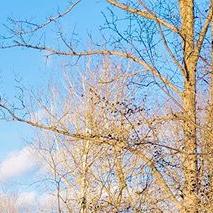}
};
\draw(3.89,15.0) node[inner sep=0pt] (img8a) {
\includegraphics[width=0.6cm,height=0.6cm]{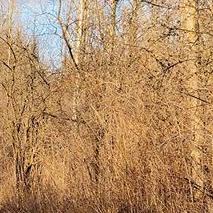}
};
\draw(3.89,14.3) node[inner sep=0pt] (img9a) {
\includegraphics[width=0.6cm,height=0.6cm]{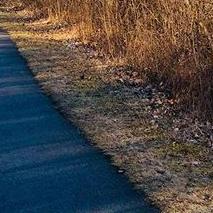}
};

\draw(0.4,13.2) node[inner sep=0pt] (img1b) {
\includegraphics[width=0.6cm,height=0.6cm]{figs/tile_0_0.jpg}
};

\draw(1.1,13.2) node[inner sep=0pt] (img2b) {
\includegraphics[width=0.6cm,height=0.6cm]{figs/tile_0_1.jpg}
};

\draw(1.8,13.2) node[inner sep=0pt] (img3b) {
\includegraphics[width=0.6cm,height=0.6cm]{figs/tile_0_2.jpg}
};

\draw(2.5,13.2) node[inner sep=0pt] (img4b) {
\includegraphics[width=0.6cm,height=0.6cm]{figs/tile_1_0.jpg}
};

\draw(3.2,13.2) node[inner sep=0pt] (img5b) {
\includegraphics[width=0.6cm,height=0.6cm]{figs/tile_1_1.jpg}
};

\draw(3.9,13.2) node[inner sep=0pt] (img6b) {
\includegraphics[width=0.6cm,height=0.6cm]{figs/tile_1_2.jpg}
};

\draw(4.6,13.2) node[inner sep=0pt] (img7b) {
\includegraphics[width=0.6cm,height=0.6cm]{figs/tile_2_0.jpg}
};

\draw(5.3,13.2) node[inner sep=0pt] (img8b) {
\includegraphics[width=0.6cm,height=0.6cm]{figs/tile_2_1.jpg}
};

\draw(6.0,13.2) node[inner sep=0pt] (img9b) {
\includegraphics[width=0.6cm,height=0.6cm]{figs/tile_2_2.jpg}
};

\draw[->, thick, inner sep=2pt] (image) -- (img2a);
\draw[->, thick, inner sep=2pt] (img6a) -- (img5b);

\path[->](3.2,12.2) node[blockMA] (proj) {
\scriptsize \textbf{Projeção Linear} 
};

\path[->](-2.3,13.8) node[inner sep=1pt] (prompt) {
\tiny \textbf{Bloco de imagem} 
};

\path[->](-2.3,13.6) node[inner sep=1pt] (prompt) {
\tiny P pixels 
};

\draw(-2.3,13.1) node[inner sep=0pt] (Itile) {
\includegraphics[width=0.6cm,height=0.6cm]{figs/tile_2_1.jpg}
};

\draw [decorate,decoration={brace,amplitude=2pt,mirror,raise=5pt}]
  (-2.1,12.8) -- (-2.1,13.4)
  node[midway,xshift=1.6em]{\tiny P pixels};

\draw [decorate,decoration={brace,amplitude=2pt,raise=5pt}]
  (-2.6,13.28) -- (-1.95,13.28)
{};

\node[draw=black, fill=black, thick, circle, minimum size=0.2cm, inner sep=0pt] at (0.12,12.2) {};
\draw[-, line width=0.8pt, dotted, inner sep=2pt] (-1.0,12.2) -- (0.12,12.2);
\node[draw=black, dotted, line width=0.8pt, fill=none, minimum width=2.7cm, minimum height=4.7cm]
   (box)  at (-2.35,11.6) {};
\path[->](-2.3,12.2) node[blockSmall,inner sep=2pt] (linear) {
\scriptsize \textbf{Linearização} 
};
\draw[->, thick, inner sep=2pt] (Itile) -- (linear);

\path[->](-2.3,11.42) node[inner sep=1pt] (info0) {
\scriptsize $P \times P$ \tiny pixels 
};

\draw[->, thick, inner sep=0pt] (linear) -- (info0);

\node[draw=black, fill=black, circle, minimum size=0.14cm, inner sep=0pt] (PxA0) at (-2.7,11.2) {};
\node[draw=black, fill=black, circle, minimum size=0.14cm, inner sep=0pt] (PxA1) at (-2.5,11.2) {};
\node[inner sep=2pt] (PxADots) at (-2.2,11.2) {\scriptsize $\dots$};
\node[draw=black, fill=black, circle, minimum size=0.14cm, inner sep=0pt] (PxAN) at (-1.95,11.2) {};

\node[draw=black, fill=black, circle, minimum size=0.14cm, inner sep=0pt] (PxB0) at (-2.9,10.5) {};
\node[draw=black, fill=black, circle, minimum size=0.14cm, inner sep=0pt] (PxB1) at (-2.7,10.5) {};
\node[draw=black, fill=black, circle, minimum size=0.14cm, inner sep=0pt] (PxB2) at (-2.5,10.5) {};
\node[inner sep=2pt] (PxBDots) at (-2.2,10.5) {\scriptsize $\dots$};
\node[draw=black, fill=black, circle, minimum size=0.14cm, inner sep=0pt] (PxBNA) at (-1.95,10.5) {};
\node[draw=black, fill=black, circle, minimum size=0.14cm, inner sep=0pt] (PxBNB) at (-1.75,10.5) {};

\draw[-, inner sep=2pt] (PxA0) -- (PxB0);
\draw[-, inner sep=2pt] (PxA0) -- (PxB1);
\draw[-, inner sep=2pt] (PxA0) -- (PxB2);
\draw[-, inner sep=2pt] (PxA0) -- (PxBDots);
\draw[-, inner sep=2pt] (PxA0) -- (PxBNA);
\draw[-, inner sep=2pt] (PxA0) -- (PxBNB);

\draw[-, inner sep=2pt] (PxA1) -- (PxB0);
\draw[-, inner sep=2pt] (PxA1) -- (PxB1);
\draw[-, inner sep=2pt] (PxA1) -- (PxB2);
\draw[-, inner sep=2pt] (PxA1) -- (PxBDots);
\draw[-, inner sep=2pt] (PxA1) -- (PxBNA);
\draw[-, inner sep=2pt] (PxA1) -- (PxBNB);

\draw[-, inner sep=2pt] (PxAN) -- (PxB0);
\draw[-, inner sep=2pt] (PxAN) -- (PxB1);
\draw[-, inner sep=2pt] (PxAN) -- (PxB2);
\draw[-, inner sep=2pt] (PxAN) -- (PxBDots);
\draw[-, inner sep=2pt] (PxAN) -- (PxBNA);
\draw[-, inner sep=2pt] (PxAN) -- (PxBNB);

\path[->](-2.9,10.15) node[inner sep=0pt] (info1) {
\scriptsize $Q$ \tiny pixels
};

\path[->](-3.25,11.05) node[inner sep=0pt] {
\scriptsize Dense 
};
\path[->](-3.25,10.75) node[inner sep=0pt] {
\scriptsize Layer 
};

\node[draw=black!50!black, fill=white!20, rounded corners=3pt, thick, minimum width=0.6cm, minimum height=0.4cm] (emb0)
    at (-2.3,9.6) {};

\draw[->, thick, inner sep=0pt] (-2.3,10.3) -- (emb0);

\draw[->,thick] (img1b.south) -- ($(img1b.south)!(proj.north)!(img1b.south)$);
\draw[->,thick] (img2b.south) -- ($(img2b.south)!(proj.north)!(img2b.south)$);
\draw[->,thick] (img3b.south) -- ($(img3b.south)!(proj.north)!(img3b.south)$);
\draw[->,thick] (img4b.south) -- ($(img4b.south)!(proj.north)!(img4b.south)$);
\draw[->,thick] (img5b.south) -- ($(img5b.south)!(proj.north)!(img5b.south)$);
\draw[->,thick] (img6b.south) -- ($(img6b.south)!(proj.north)!(img6b.south)$);
\draw[->,thick] (img7b.south) -- ($(img7b.south)!(proj.north)!(img7b.south)$);
\draw[->,thick] (img8b.south) -- ($(img8b.south)!(proj.north)!(img8b.south)$);
\draw[->,thick] (img9b.south) -- ($(img9b.south)!(proj.north)!(img9b.south)$);

\node[draw=black!50!black, fill=white!20, rounded corners=3pt, thick, minimum width=0.6cm, minimum height=0.4cm, inner sep=0pt] (text0)
    at (-0.3,11.3) {\tiny CLS};

\node[draw=black!50!black, fill=white!20, rounded corners=3pt, thick, minimum width=0.6cm, minimum height=0.4cm] (text1)
    at (0.4,11.3) {};

\node[draw=black!50!black, fill=white!20, rounded corners=3pt, thick, minimum width=0.6cm, minimum height=0.4cm] (text2)
    at (1.1,11.3) {};

\node[draw=black!50!black, fill=white!20, rounded corners=3pt, thick, minimum width=0.6cm, minimum height=0.4cm] (text3)
    at (1.8,11.3) {};    

\node[draw=black!50!black, fill=white!20, rounded corners=3pt, thick, minimum width=0.6cm, minimum height=0.4cm] (text4)
    at (2.5,11.3) {}; 

\node[draw=black!50!black, fill=white!20, rounded corners=3pt, thick, minimum width=0.6cm, minimum height=0.4cm] (text5)
    at (3.2,11.3) {};

\node[draw=black!50!black, fill=white!20, rounded corners=3pt, thick, minimum width=0.6cm, minimum height=0.4cm] (text6)
    at (3.9,11.3) {};
    
\node[draw=black!50!black, fill=white!20, rounded corners=3pt, thick, minimum width=0.6cm, minimum height=0.4cm] (text7)
    at (4.6,11.3) {};

\node[draw=black!50!black, fill=white!20, rounded corners=3pt, thick, minimum width=0.6cm, minimum height=0.4cm] (text8)
    at (5.3,11.3) {};

\node[draw=black!50!black, fill=white!20, rounded corners=3pt, thick, minimum width=0.6cm, minimum height=0.4cm] (text9)
    at (6.0,11.3) {};    
    
\draw[<-,thick] (text1.north) -- ($(text1.north)!(proj.south)!(text1.north)$);
\draw[<-,thick] (text2.north) -- ($(text2.north)!(proj.south)!(text2.north)$);
\draw[<-,thick] (text3.north) -- ($(text3.north)!(proj.south)!(text3.north)$);
\draw[<-,thick] (text4.north) -- ($(text4.north)!(proj.south)!(text4.north)$);
\draw[<-,thick] (text5.north) -- ($(text5.north)!(proj.south)!(text5.north)$);
\draw[<-,thick] (text6.north) -- ($(text6.north)!(proj.south)!(text6.north)$);
\draw[<-,thick] (text7.north) -- ($(text7.north)!(proj.south)!(text7.north)$);
\draw[<-,thick] (text8.north) -- ($(text8.north)!(proj.south)!(text8.north)$);
\draw[<-,thick] (text9.north) -- ($(text9.north)!(proj.south)!(text9.north)$);

\path[->](2.85,10.45) node[blockMB] (tencoder) {
\scriptsize \textbf{Visual Transformer} 
};

\draw[->,thick] (text0.south) -- ($(text0.south)!(tencoder.north)!(text0.south)$);
\draw[->,thick] (text1.south) -- ($(text1.south)!(tencoder.north)!(text1.south)$);
\draw[->,thick] (text2.south) -- ($(text2.south)!(tencoder.north)!(text2.south)$);
\draw[->,thick] (text3.south) -- ($(text3.south)!(tencoder.north)!(text3.south)$);
\draw[->,thick] (text4.south) -- ($(text4.south)!(tencoder.north)!(text4.south)$);
\draw[->,thick] (text5.south) -- ($(text5.south)!(tencoder.north)!(text5.south)$);
\draw[->,thick] (text6.south) -- ($(text6.south)!(tencoder.north)!(text6.south)$);
\draw[->,thick] (text7.south) -- ($(text7.south)!(tencoder.north)!(text7.south)$);
\draw[->,thick] (text8.south) -- ($(text8.south)!(tencoder.north)!(text8.south)$);
\draw[->,thick] (text9.south) -- ($(text9.south)!(tencoder.north)!(text9.south)$);

\node[draw=black!50!black, fill=red!20, rounded corners=3pt, thick, minimum width=0.6cm, minimum height=0.4cm, inner sep=0pt] (timg0)
    at (-0.3,9.6) {};

\node[draw=black!50!black, fill=red!20, rounded corners=3pt, thick, minimum width=0.6cm, minimum height=0.4cm] (timg1)
    at (0.4,9.6) {};

\node[draw=black!50!black, fill=red!20, rounded corners=3pt, thick, minimum width=0.6cm, minimum height=0.4cm] (timg2)
    at (1.1,9.6) {};

\node[draw=black!50!black, fill=red!20, rounded corners=3pt, thick, minimum width=0.6cm, minimum height=0.4cm] (timg3)
    at (1.8,9.6) {};

\node[draw=black!50!black, fill=red!20, rounded corners=3pt, thick, minimum width=0.6cm, minimum height=0.4cm] (timg4)
    at (2.5,9.6) {};

\node[draw=black!50!black, fill=red!20, rounded corners=3pt, thick, minimum width=0.6cm, minimum height=0.4cm] (timg5)
    at (3.2,9.6) {};

\node[draw=black!50!black, fill=red!20, rounded corners=3pt, thick, minimum width=0.6cm, minimum height=0.4cm] (timg6)
    at (3.9,9.6) {};

\node[draw=black!50!black, fill=red!20, rounded corners=3pt, thick, minimum width=0.6cm, minimum height=0.4cm] (timg7)
    at (4.6,9.6) {};

\node[draw=black!50!black, fill=red!20, rounded corners=3pt, thick, minimum width=0.6cm, minimum height=0.4cm] (timg8)
    at (5.3,9.6) {};

\node[draw=black!50!black, fill=red!20, rounded corners=3pt, thick, minimum width=0.6cm, minimum height=0.4cm] (timg9)
    at (6.0,9.6) {};    

\draw[<-,thick] (timg0.north) -- ($(timg0.north)!(tencoder.south)!(timg0.north)$);
\draw[<-,thick] (timg1.north) -- ($(timg1.north)!(tencoder.south)!(timg1.north)$);
\draw[<-,thick] (timg2.north) -- ($(timg2.north)!(tencoder.south)!(timg2.north)$);
\draw[<-,thick] (timg3.north) -- ($(timg3.north)!(tencoder.south)!(timg3.north)$);
\draw[<-,thick] (timg4.north) -- ($(timg4.north)!(tencoder.south)!(timg4.north)$);
\draw[<-,thick] (timg5.north) -- ($(timg5.north)!(tencoder.south)!(timg5.north)$);
\draw[<-,thick] (timg6.north) -- ($(timg6.north)!(tencoder.south)!(timg6.north)$);
\draw[<-,thick] (timg7.north) -- ($(timg7.north)!(tencoder.south)!(timg7.north)$);
\draw[<-,thick] (timg8.north) -- ($(timg8.north)!(tencoder.south)!(timg8.north)$);
\draw[<-,thick] (timg9.north) -- ($(timg9.north)!(tencoder.south)!(timg9.north)$);


\path[->,color=blue](9.5,13.6) node[inner sep=1pt] (prompt) {
\footnotesize \textbf{Prompt} 
};

\path[->](9.5,13.05) node[inner sep=1pt] (prompt) {
\scriptsize \textbf{"Descreva a imagem"} 
};

\path[->](8.5,11.3) node[inner sep=2pt] (prompt1) {
\scriptsize \textbf{`Descreva'} 
};

\path[->](9.5,11.3) node[inner sep=2pt] (prompt2) {
\scriptsize \textbf{`a'} 
};

\path[->](10.5,11.25) node[inner sep=2pt] (prompt3) {
\scriptsize \textbf{`imagem'} 
};

\path[->](9.5,12.2) node[blockTA] (tokenizer) {
\scriptsize \textbf{Tokenizador} 
};

\path[->](9.5,10.45) node[blockTA] (embedding) {
\scriptsize \textbf{Embedding} 
};

\draw[->,thick] (prompt) -- (tokenizer);
\draw[<-,thick] (prompt1.north) -- ($(prompt1.north)!(tokenizer.south)!(prompt1.north)$);
\draw[<-,thick] (prompt2.north) -- ($(prompt2.north)!(tokenizer.south)!(prompt2.north)$);
\draw[<-,thick] (prompt3.north) -- ($(prompt3.north)!(tokenizer.south)!(prompt3.north)$);

\draw[->,thick] (prompt1.south) -- ($(prompt1.south)!(embedding.north)!(prompt1.south)$);
\draw[->,thick] (prompt2.south) -- ($(prompt2.south)!(embedding.north)!(prompt2.south)$);
\draw[->,thick] (prompt3.south) -- ($(prompt3.south)!(embedding.north)!(prompt3.south)$);

\node[draw=black!50!black, fill=blue!20, rounded corners=3pt, thick, minimum width=0.6cm, minimum height=0.4cm] (Ttext1)
    at (8.5,9.6) {};

\node[draw=black!50!black, fill=blue!20, rounded corners=3pt, thick, minimum width=0.6cm, minimum height=0.4cm] (Ttext2)
    at (9.5,9.6) {};    

\node[draw=black!50!black, fill=blue!20, rounded corners=3pt, thick, minimum width=0.6cm, minimum height=0.4cm] (Ttext3)
    at (10.5,9.6) {};  

\draw[<-,thick] (Ttext1.north) -- ($(Ttext1.north)!(embedding.south)!(Ttext1.north)$);
\draw[<-,thick] (Ttext2.north) -- ($(Ttext2.north)!(embedding.south)!(Ttext2.north)$);
\draw[<-,thick] (Ttext3.north) -- ($(Ttext3.north)!(embedding.south)!(Ttext3.north)$);

\path[->](3.0,8.5) node[inner sep=1pt] (prompt) {
\small\textbf{(a) Encoder de imagem} 
};

\path[->](9.5,8.5) node[inner sep=1pt] (prompt) {
\small\textbf{(b) Encoder de texto} 
};

\end{tikzpicture}

%% file: scheme-modalities.tex
\begin{tikzpicture}[scale=0.85, every node/.style={scale=0.85}]

\tikzset{blockNA/.style={draw, rectangle, text centered, drop shadow, fill=gray!20!white, text width=3.7cm, minimum height=2.9cm, inner sep=2pt}}

\tikzset{blockNB/.style={draw, rectangle, text centered, drop shadow, fill=white!20!white, text width=2.9cm, minimum height=0.4cm, inner sep=2pt}}

\draw(0.05,15.4) node[inner sep=1pt] (image) {
\includegraphics[width=2.0cm,height=2.0cm]{figs/DeepImage.jpg}
};

\node[draw=black!50!black, fill=red!20, rounded corners=3pt, thick, minimum width=0.6cm, minimum height=0.4cm] (cx0)
    at (-0.9,13.3) {};  

\node[draw=black!50!black, fill=red!20, rounded corners=3pt, thick, minimum width=0.6cm, minimum height=0.4cm] (cx1)
    at (-0.2,13.3) {};

\path[->](0.4,13.3) node[] (cx2) {
\normalsize \textbf{...} 
};    

\node[draw=black!50!black, fill=red!20, rounded corners=3pt, thick, minimum width=0.6cm, minimum height=0.4cm] (cxn)
    at (+1.0,13.3) {};    
    
\path[->](2.7,13.8) node[inner sep=1pt,color=blue] (EncT) {
\scriptsize \textbf{Encoder de texto} 
};

\path[->](0.0,13.8) node[inner sep=1pt,color=red] (EncI) {
\scriptsize \textbf{Encoder de imagem} 
};

\path[->,color=blue](2.7,14.9) node[inner sep=1pt] (prompt0) {
\footnotesize \textbf{Prompt} 
};

\path[->](2.7,14.5) node[inner sep=1pt] (prompt) {
\scriptsize \textbf{"Descreva a imagem"} 
};

\node[draw=black!50!black, fill=blue!20, rounded corners=3pt, thick, minimum width=0.6cm, minimum height=0.4cm] (timg0)
    at (2.0,13.3) {};  

\node[draw=black!50!black, fill=blue!20, rounded corners=3pt, thick, minimum width=0.6cm, minimum height=0.4cm] (timg1)
    at (2.7,13.3) {};     

\node[draw=black!50!black, fill=blue!20, rounded corners=3pt, thick, minimum width=0.6cm, minimum height=0.4cm] (timg2)
    at (3.4,13.3) {}; 

\path[->](1.3,11.15) node[blockNA] (MLLM) {

};

\path[->](1.3,11.8) node[blockNB] (LN) {
\scriptsize \textbf{...} 
};

\path[->](1.3,11.4) node[blockNB] (LN) {
\scriptsize \textbf{Layer Norm} 
};

\path[->](1.3,11.0) node[blockNB] (MHA) {
\scriptsize \textbf{Multi-Head Attention} 
};

\path[->](1.3,10.6) node[blockNB] (MHA) {
\scriptsize \textbf{Dropout} 
};

\path[->](1.3,10.2) node[blockNB] (LN) {
\scriptsize \textbf{...} 
};

\path[->](2.0,12.3) node[] (MLLM) {
\footnotesize \textbf{MLLM} 
};

\node[draw=red, line width=0.5pt, fill=none, minimum width=2.8cm, minimum height=0.7cm]
   (box1)  at (0.05,13.3) {};

\node[draw=blue, line width=0.5pt, fill=none, minimum width=2.2cm, minimum height=0.7cm]
   (box2)  at (2.7,13.3) {};

\draw[->,blue,thick] (prompt.south) -- ($(prompt.south)!(EncT.north)!(prompt.south)$);
\draw[->,red,thick] (image.south) -- ($(image.south)!(EncI.north)!(image.south)$);

\draw[->,blue,thick] (box2.south) -- (2.7,12.6);
\draw[->,red,thick] (box1.south) -- (0.05,12.6);

\path[->](1.3,9.3) node[] (MLLM) {
\small\textbf{(a) Fusão precoce} 
};

\end{tikzpicture} \hspace{8pt}
\begin{tikzpicture}[scale=0.85, every node/.style={scale=0.85}]

\tikzset{blockNA/.style={draw, rectangle, text centered, drop shadow, fill=gray!20!white, text width=3.7cm, minimum height=2.9cm, inner sep=2pt}}

\tikzset{blockNB/.style={draw, rectangle, text centered, drop shadow, fill=white!20!white, text width=2.9cm, minimum height=0.4cm, inner sep=2pt}}

\draw(0.05,15.4) node[inner sep=1pt] (image) {
\includegraphics[width=2.0cm,height=2.0cm]{figs/DeepImage.jpg}
};

\node[draw=black!50!black, fill=red!20, rounded corners=3pt, thick, minimum width=0.6cm, minimum height=0.4cm] (cx0)
    at (-0.9,13.3) {};  

\node[draw=black!50!black, fill=red!20, rounded corners=3pt, thick, minimum width=0.6cm, minimum height=0.4cm] (cx1)
    at (-0.2,13.3) {};

\path[->](0.4,13.3) node[] (cx2) {
\normalsize \textbf{...} 
};    

\path[->](2.7,13.8) node[inner sep=1pt,color=blue] (EncT) {
\scriptsize \textbf{Encoder de texto} 
};

\path[->](0.0,13.8) node[inner sep=1pt,color=red] (EncI) {
\scriptsize \textbf{Encoder de imagem} 
};

\node[draw=black!50!black, fill=red!20, rounded corners=3pt, thick, minimum width=0.6cm, minimum height=0.4cm] (cxn)
    at (+1.0,13.3) {};

\path[->,color=blue](2.7,14.9) node[inner sep=1pt] (prompt0) {
\footnotesize \textbf{Prompt} 
};

\path[->](2.7,14.5) node[inner sep=1pt] (prompt) {
\scriptsize \textbf{"Descreva a imagem"} 
};

\node[draw=black!50!black, fill=blue!20, rounded corners=3pt, thick, minimum width=0.6cm, minimum height=0.4cm] (timg0)
    at (2.0,13.3) {};  

\node[draw=black!50!black, fill=blue!20, rounded corners=3pt, thick, minimum width=0.6cm, minimum height=0.4cm] (timg1)
    at (2.7,13.3) {};     

\node[draw=black!50!black, fill=blue!20, rounded corners=3pt, thick, minimum width=0.6cm, minimum height=0.4cm] (timg2)
    at (3.4,13.3) {}; 

\path[->](1.3,11.15) node[blockNA] (MLLM) {

};

\path[->](1.3,11.8) node[blockNB] (LN) {
\scriptsize \textbf{...} 
};

\path[->](1.3,11.4) node[blockNB] (LN) {
\scriptsize \textbf{Layer Norm} 
};

\path[->](1.3,11.0) node[blockNB,fill=red!50!white] (MHA) {
\scriptsize \textbf{Multi-Head Attention} 
};

\path[->](1.3,10.6) node[blockNB] (MHA) {
\scriptsize \textbf{Dropout} 
};

\path[->](1.3,10.2) node[blockNB] (LN) {
\scriptsize \textbf{...} 
};

\path[->](2.0,12.3) node[] (MLLM) {
\footnotesize \textbf{MLLM} 
};

\node[draw=red, line width=0.5pt, fill=none, minimum width=2.8cm, minimum height=0.7cm]
   (box1)  at (0.05,13.3) {};

\node[draw=blue, line width=0.5pt, fill=none, minimum width=2.2cm, minimum height=0.7cm]
   (box2)  at (2.7,13.3) {};

\draw[->,blue,thick] (prompt.south) -- ($(prompt.south)!(EncT.north)!(prompt.south)$);
\draw[->,red,thick] (image.south) -- ($(image.south)!(EncI.north)!(image.south)$);

\draw[->,blue,thick] (box2.south) -- (2.7,12.6);
\draw[-,red,thick] (-1.0,12.95) -- (-1.0,11.0);
\draw[->,red,thick] (-1.0,11.0) -- (-0.3,11.0);

\path[->](1.3,9.3) node[] (MLLM) {
\small\textbf{(b) Fusão tardia.}
};

\end{tikzpicture}
\hspace{8pt}
\begin{tikzpicture}[scale=0.85, every node/.style={scale=0.85}]

\tikzset{blockNA/.style={draw, rectangle, text centered, drop shadow, fill=gray!20!white, text width=3.7cm, minimum height=2.9cm, inner sep=2pt}}

\tikzset{blockNB/.style={draw, rectangle, text centered, drop shadow, fill=white!20!white, text width=2.9cm, minimum height=0.4cm, inner sep=2pt}}

\draw(0.05,15.4) node[inner sep=1pt] (image) {
\includegraphics[width=2.0cm,height=2.0cm]{figs/DeepImage.jpg}
};

\node[draw=black!50!black, fill=red!20, rounded corners=3pt, thick, minimum width=0.6cm, minimum height=0.4cm] (cx0)
    at (-0.9,13.3) {};  

\node[draw=black!50!black, fill=red!20, rounded corners=3pt, thick, minimum width=0.6cm, minimum height=0.4cm] (cx1)
    at (-0.2,13.3) {};

\path[->](0.4,13.3) node[] (cx2) {
\normalsize \textbf{...} 
};    

\node[draw=black!50!black, fill=red!20, rounded corners=3pt, thick, minimum width=0.6cm, minimum height=0.4cm] (cxn)
    at (+1.0,13.3) {};

\path[->,color=blue](2.7,14.9) node[inner sep=1pt] (prompt0) {
\footnotesize \textbf{Prompt} 
};

\path[->](2.7,14.5) node[inner sep=1pt] (prompt) {
\scriptsize \textbf{"Descreva a imagem"} 
};

\path[->](2.7,13.8) node[inner sep=1pt,color=blue] (EncT) {
\scriptsize \textbf{Encoder de texto} 
};

\path[->](0.0,13.8) node[inner sep=1pt,color=red] (EncI) {
\scriptsize \textbf{Encoder de imagem} 
};

\node[draw=black!50!black, fill=blue!20, rounded corners=3pt, thick, minimum width=0.6cm, minimum height=0.4cm] (timg0)
    at (2.0,13.3) {};  

\node[draw=black!50!black, fill=blue!20, rounded corners=3pt, thick, minimum width=0.6cm, minimum height=0.4cm] (timg1)
    at (2.7,13.3) {};     

\node[draw=black!50!black, fill=blue!20, rounded corners=3pt, thick, minimum width=0.6cm, minimum height=0.4cm] (timg2)
    at (3.4,13.3) {}; 

\path[->](1.3,11.15) node[blockNA] (MLLM) {

};

\path[->](1.3,11.8) node[blockNB] (LN) {
\scriptsize \textbf{...} 
};

\path[->](1.3,11.4) node[blockNB] (LN) {
\scriptsize \textbf{Layer Norm} 
};

\path[->](1.3,11.0) node[blockNB,fill=red!50!white] (MHA) {
\scriptsize \textbf{Multi-Head Attention} 
};

\path[->](1.3,10.6) node[blockNB] (MHA) {
\scriptsize \textbf{Dropout} 
};

\path[->](1.3,10.2) node[blockNB] (LN) {
\scriptsize \textbf{...} 
};

\path[->](2.0,12.3) node[] (MLLM) {
\footnotesize \textbf{MLLM} 
};

\node[draw=red, line width=0.5pt, fill=none, minimum width=2.8cm, minimum height=0.7cm]
   (box1)  at (0.05,13.3) {};

\node[draw=blue, line width=0.5pt, fill=none, minimum width=2.2cm, minimum height=0.7cm]
   (box2)  at (2.7,13.3) {};

\draw[->,blue,thick] (prompt.south) -- ($(prompt.south)!(EncT.north)!(prompt.south)$);
\draw[->,red,thick] (image.south) -- ($(image.south)!(EncI.north)!(image.south)$);

\draw[->,blue,thick] (box2.south) -- (2.7,12.6);
\draw[-,red,thick] (-1.0,12.95) -- (-1.0,11.0);
\draw[->,red,thick] (-1.0,11.0) -- (-0.3,11.0);

\draw[->,red,thick] (box1.south) -- (0.05,12.6);

\node[draw=black, dashed, thick, circle, minimum size=0.7cm] at (0.05,12.8) {};
\node[draw=black, dashed, thick, circle, minimum size=0.7cm] at (-0.5,11.0) {};

\path[->](1.3,9.3) node[] (MLLM) {
\small\textbf{(c) Híbrida} 
};

\end{tikzpicture}

%% file: texFigures/captionsB.tex
\newcommand{\placeimage}[5]{%
  
{
  \vspace{-2pt}
  \centering
  {\scriptsize \textbf{#1}} \\[3pt] 
  \includegraphics[width=4.0cm,height=4.0cm]{#2}
}  

  {\fontsize{9.5pt}{10pt}\selectfont
  \textbf{\underline{\VLMos}:} \justifying ``\textit{#3}''

  \vspace{6pt}

  \noindent \textbf{\underline{\VLMoai}:} ``\textit{#4}''

  \vspace{6pt}

  \noindent \textbf{\underline{\VLMds}:} ``\textit{#5}''

  \vspace{2pt}
  }
}
\noindent\hrulefill \vspace{3pt}
\placeimage{Imagem de Entrada para os MLLMs}{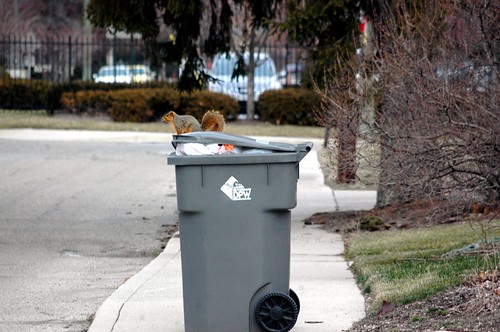}{A imagem mostra uma lata de lixo cinza na calçada, junto ao meio-fio. A lata tem uma tampa e parece estar cheia de lixo. Não há árvores ou outros objetos na imagem, apenas a calçada e uma rua com carros circulando. O céu está nublado e encoberto.}{A imagem mostra uma lixeira cinza posicionada na calçada, com alguns esquilos empoleirados na borda. Os esquilos parecem estar procurando comida, com seus rabos erguidos. O fundo inclui uma cena de bairro com árvores, alguns arbustos secos e uma vista distante de carros estacionados além de uma cerca preta. O pavimento é de concreto claro e a grama ao lado da calçada parece acastanhada, possivelmente indicando uma mudança de estação. A atmosfera parece calma e residencial, típica de uma área suburbana.}{A imagem mostra uma lata de lixo cinza em uma calçada. A lata está aberta e um esquilo está em pé na borda. A lata está localizada na calçada, com uma calçada e uma cerca ao fundo.}
\noindent\hrulefill